\documentclass{article} %
\usepackage{iclr2027_conference,times}

\usepackage{amsmath,amsfonts,bm}

\def\eqref#1{equation~\ref{#1}}

\def\1{\bm{1}}

\DeclareMathAlphabet{\mathsfit}{\encodingdefault}{\sfdefault}{m}{sl}
\SetMathAlphabet{\mathsfit}{bold}{\encodingdefault}{\sfdefault}{bx}{n}

\usepackage{graphicx}
\usepackage{hyperref}
\makeatletter
\ifdefined\pdfrunninglinkoff
  \let\pdfannot@link@off@@\pdfrunninglinkoff
  \let\pdfannot@link@on@@\pdfrunninglinkon
\fi
\makeatother
\usepackage{url}
\usepackage{booktabs}
\usepackage{multirow}
\usepackage{wrapfig}
\usepackage{float}
\usepackage{needspace}
\usepackage{amsmath,amssymb}
\usepackage{algorithm}
\usepackage{algpseudocode}

\algrenewcommand{\algorithmicrequire}{\textbf{Require:}}
\algrenewcommand{\algorithmicensure}{\textbf{Ensure:}}

\algrenewcommand{\alglinenumber}[1]{\normalfont #1:}

\title{SemPSG: A Semantic Channel-Aware Foundation Model for Polysomnography Analysis}

\author{
\rule{0pt}{2.4em}%
\begin{minipage}[t]{0.97\textwidth}
\raggedright
\textbf{Junyu Chen}$^{1}$, \textbf{Chenxi Liu}$^{1}$\thanks{Corresponding author.}~~,~\textbf{Shiqin Tang}$^{1}$, \textbf{Hao Miao}$^{2}$, \textbf{Wanyun Ling}$^{3}$, \textbf{Ziyue Li}$^{3}$,\\
\textbf{Hongbin Liu}$^{1}$, \textbf{Gaofeng Meng}$^{1}$\\[0.35em]
\normalfont\small
$^{1}$Center for Artificial Intelligence and Robotics, Hong Kong Institute of Science \& Innovation, CAS\\
$^{2}$Shenzhen Institute for Advanced Study, University of Electronic Science and Technology of China\\
$^{3}$Department of Operations and Technology, Technical University of Munich\\
\footnotesize
\resizebox{\linewidth}{!}{%
\texttt{\{junyu.chen, chenxi.liu, shiqin.tang, hongbin.liu, gaofeng.meng\}@cair-cas.hk.org}%
}\\[0.15em]
\mbox{\texttt{hao-miao@outlook.com, \{wanyun.ling, ziyue.li\}@tum.de}}
\end{minipage}
}

\iclrfinalcopy %
\begin{document}

\maketitle
\fancyhead{}
\renewcommand{\headrulewidth}{0pt}

\begin{abstract}
Polysomnography (PSG) integrates multiple physiological signals to provide a comprehensive characterization of human sleep. The growing availability of large-scale PSG recordings creates an opportunity to develop unified time series models across diverse applications. Existing studies often overlook the structured semantics encoded in heterogeneous PSG channel identities. Meanwhile, multi-scale temporal dynamics and entangled intra- and inter-modality dependencies pose challenges for unified PSG representation learning.
In this paper, we propose \textbf{SemPSG}, a \underline{Sem}antic channel-aware foundation model for heterogeneous \underline{PSG} analysis. SemPSG learns representations from time series and image views
while incorporating the channel semantics. Notably, it introduces a channel semantics tokenizer that encodes physiological information from channel identities. A semantic-aware time-series and image architecture jointly captures long-range temporal dynamics, time-frequency patterns, and local waveform morphology, alleviating the trade-off across temporal scales. Finally, the intra- and inter-modality encoders explicitly model channel-level dependencies within each modality and cross physiological interactions.
We evaluate SemPSG on six sleep and health-related tasks, experimental results demonstrate SemPSG outperforms general-purpose and PSG-specific foundation models, achieving a 9.6 percentage-point gain for sleep-disordered breathing classification, while demonstrating transferability across heterogeneous PSG and channel configurations.
\end{abstract}

\section{Introduction}
Polysomnography (PSG) provides a comprehensive view of human sleep by continuously capturing multiple physiological signals, including electroencephalography (EEG), electrooculography (EOG), electromyography (EMG), electrocardiography (ECG), and respiratory signals~\citep{Troester2023AASM, lee2025explainable}. Jointly, these signals capture brain activity, eye movements, muscle activity, cardiac activity, and respiratory dynamics~\citep{krugliakova2026}, respectively. The synchronized recordings form multi-physiological time series that support diverse downstream analyses~\citep{ma2025sleepsmc}.
However, conventional PSG models are commonly developed for individual tasks,  such as sleep staging or event detection, and rely on task-specific annotations~\citep{phan2021xsleepnet, wang2024generalizable, kjaer2026expert}.
The growing availability of large-scale PSG recordings creates an opportunity to develop unified time series models that learn reusable physiological representations and transfer them across diverse applications~\citep{thapa2024sleepfm, shuai2026osf}.

\begin{figure}[t]
    \centering
    \vspace{-0.1cm}
    \includegraphics[width=1.0\textwidth]{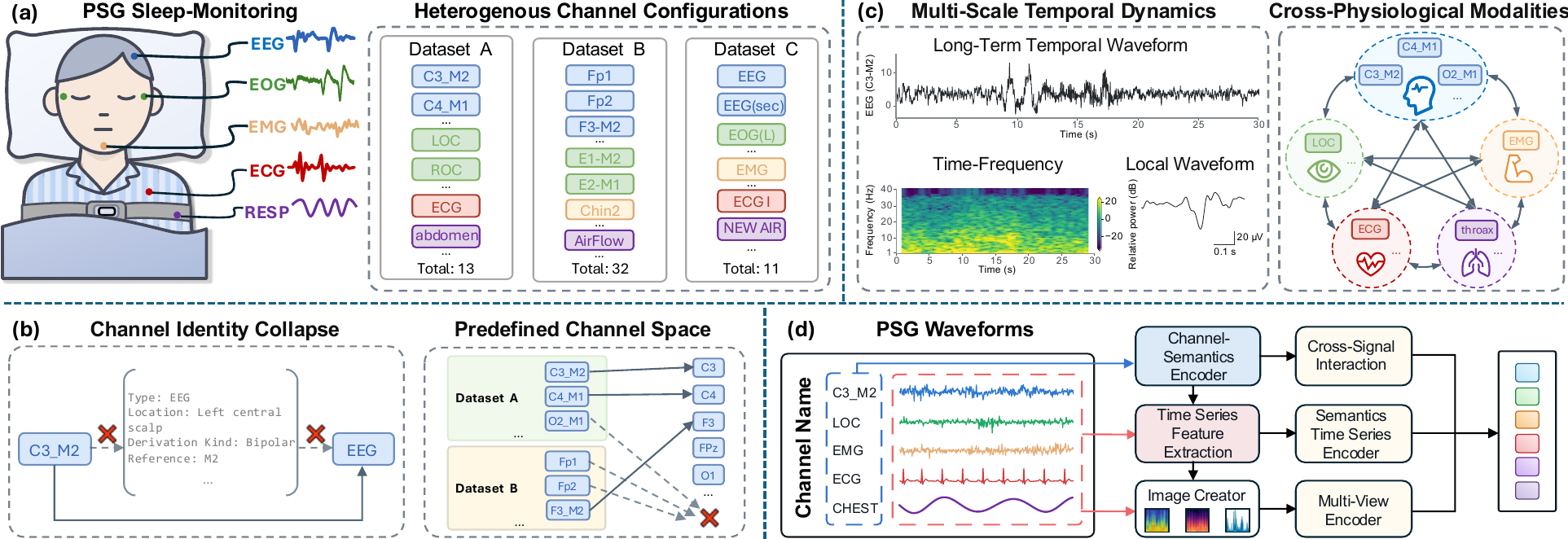}
    \vspace{-0.4cm}
    \caption{\textbf{Motivation.} (a) An example of PSG monitoring. %
    (b) Limitations of conventional PSG. (c) PSG data characteristics. (d) The pipeline of SemPSG.}
    \vspace{-0.7cm}
    \label{intro_fig}
\end{figure}

Recent progress in time series foundation models has demonstrated the effectiveness of self-supervised pretraining for learning transferable representations across domains~\citep{nie2023a, liu2024timer,fu2026zeus}. However, PSG differs from conventional time series. As illustrated in Figure~\ref{intro_fig} (a), PSG comprises multiple physiological modalities with distinct sleep-related temporal dynamics and complex interactions across physiological systems, making direct adaptation of general-purpose foundation models non-trivial. These characteristics have motivated the development of foundation models specifically tailored to sleep- related physiological data~\citep{xu2026sleeplm}. For example, SleepFM explores multimodal contrastive pretraining across physiological signals~\citep{thapa2026multimodal}, while SleepMaMi employs hierarchical temporal representation learning~\citep{park2026sleepmami}. Other studies investigate cross-modal alignment across heterogeneous nocturnal biosignals~\citep{yuan2026sleep2vec} and unified time-frequency modeling~\citep{huang2026unified}.

 Despite these advances, existing approaches still struggle to learn generalizable representations across heterogeneous PSG recordings due to several challenges. The first challenge lies in \textit{channel semantics heterogeneity}. PSG channels vary substantially across datasets from different acquisition systems, as illustrated in Figure~\ref{intro_fig} (a). One major issue is channel identity collapse, where distinct channels within the same physiological modality become indistinguishable when represented only by coarse modality labels, as shown in the left panel of Figure~\ref{intro_fig} (b). For example, \textit{C3-M2} is not merely an EEG channel: \textit{C3} specifies the recording location over the left central scalp region, whereas \textit{M2} specifies the reference electrode over the right mastoid. Moreover, \textit{C3-M2} and \textit{M2-C3} contain the same electrode identities but opposite derivation directions, resulting in reversed waveform. 
 Furthermore, predefined channel spaces preserve identities by assigning them to a fixed vocabulary of canonical channels, as shown in the right part of Figure~\ref{intro_fig} (b). It is difficult to accommodate unseen channels arising from different derivation configurations and naming conventions.

 The second challenge is \textit{temporal dynamics trade-off}. As shown in the left panel of Figure~\ref{intro_fig} (c), PSG signals exhibit distinct temporal dynamics at multiple scales: long-term temporal waveform characterizes the evolution of physiological states over extended periods, whereas time-frequency patterns reveal spectral characteristics; Meanwhile, local waveform morphology captures transient physiological events. These temporal dynamics provide complementary information for sleep analysis, but remains difficulties to balance for effective temporal representations.
 The third challenge is \textit{intra- and inter-modality dependency entanglement}. As illustrated in the right part of Figure~\ref{intro_fig} (c), PSG contains dependencies both within and across physiological modalities. Existing approaches often emphasize inter-modality interactions while underexploring intra-modality channel relationships~\citep{thapa2026multimodal,huang2026unified,park2026sleepmami}, causing channel-level correlations to become entangled with modality-level interactions.

 In this paper, we propose \textbf{SemPSG}, a \underline{Sem}antic channel-aware foundation model for \underline{PSG} analysis. As illustrated in Figure~\ref{intro_fig} (d), SemPSG learns representations from time series and image views while incorporating the channel semantics. 
 To address the channel semantic heterogeneity, a channel semantics tokenizer is designed to represent each channel through the structured physiological semantics encoded in its identity. These semantic representations condition both waveform feature learning and channel aggregation, preserving channel-specific information without collapsing channels into coarse modality labels and restricting them to a predefined channel space.
 To alleviate the temporal dynamics trade-off, the semantic-aware time-series module captures long-range temporal dynamics from physiological waveforms, while the time-frequency and morphology branch extracts spectral patterns and local waveform morphology through  Wavelet Transform (CWT) scalograms, Short-Time Fourier Transform (STFT) spectrograms, and morphology envelopes. The three-category views are then aggregated using learned sample-dependent weights.
 To disentangle the intra- and inter-modality dependencies, an intra-modality encoder first captures relationships among channels within each physiological modality, followed by semantic channel aggregation that forms modality-level representations. An inter-modality encoder then models interactions across different physiological systems, preventing channel-level correlations from being directly mixed with modality-level dependencies.

Overall, our main contributions are summarized as follows:

{\bf (i)} We propose \textbf{SemPSG}, a \underline{Sem}antic channel-aware foundation model for heterogeneous \underline{PSG} analysis. SemPSG introduces a channel semantics tokenizer that transforms dataset-specific channel identities into structured physiological representations, establishing a unified semantic space for knowledge sharing across heterogeneous channel configurations and acquisition systems.

{\bf (ii)} SemPSG learns a unified representation from semantic time series and image views. The semantic-aware time series module hierarchically models intra- and inter-modality dependencies, disentangling channel-level correlations from modality-level interactions, while the multi-scale temporal module balances the long-range dynamics, spectral, and local waveform patterns.

{\bf (iii)} Experiments conducted on six downstream tasks covering diverse sleep and health outcomes. SemPSG consistently outperforms general-purpose and PSG-specific foundation models, achieving a 9.6 percentage-point gain for sleep-disordered breathing classification, while demonstrating transferability across heterogeneous PSG datasets and channel configurations.

\section{Methodology}
\label{sec:method}

\begin{figure}[t]
    \centering
    \includegraphics[width=1.0\textwidth]{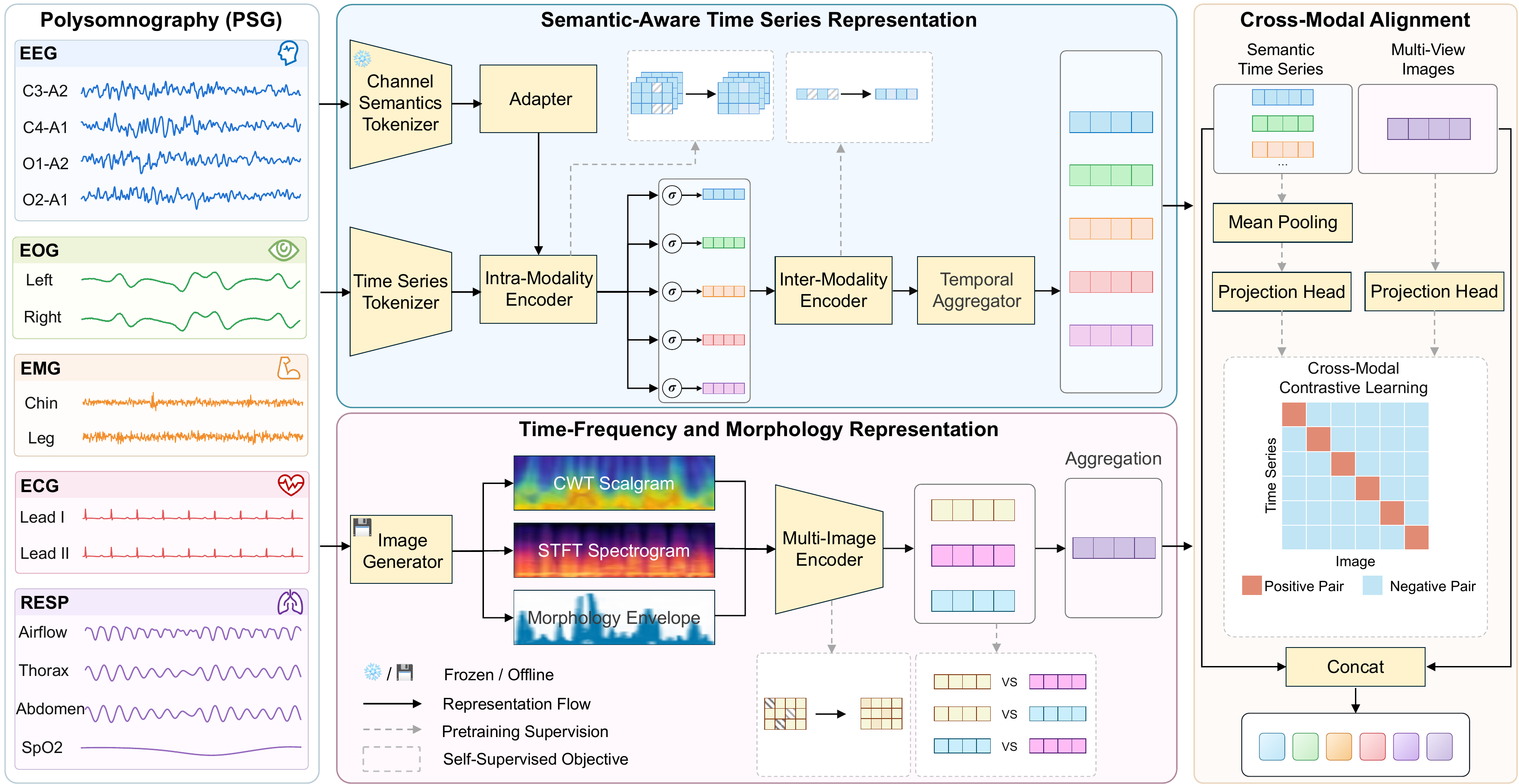}
    \vspace{-0.6cm}
    \caption{\textbf{Overall Framework of SemPSG.}}
    \label{fig:framework}
\end{figure}

SemPSG learns unified PSG representations from two views: a semantic-aware time series view and a time-frequency and morphology view, as shown in Figure~\ref{fig:framework}. In the semantic time-series view, PSG signals are tokenized into temporal representations, while channel identities are encoded into semantic tokens that capture the structured information associated with each PSG channel. The fused semantic time-series embeddings are learned with intra- and inter-modality reconstruction objectives used during pretraining. In the image view, each physiological signal is transformed into Continuous Wavelet Transform (CWT) scalograms, Short-Time Fourier Transform (STFT) spectrograms, and morphology envelopes to capture complementary time-frequency and waveform characteristics.
Finally, semantic, intra- and inter-modal, and multi-scale temporal representations are aligned in a shared space, enabling the model to capture both the semantics and physiological patterns of PSG.

\subsection{Semantic-Aware Time-Series Encoding}

To incorporate channel identity into physiological dependency modeling, the time-series branch combines channel-semantic conditioning with successive intra- and inter-modality encoding. It maps the waveforms $\mathbf{X}^{(t)}$, channel names $\mathcal{N}^{(t)}$, and validity mask $A^{(t)}$ to window-level representations $\{ \mathbf{u}^{(t)}_m\}^M_{m=1}$, retaining a separate output token for each physiological modality.

\textbf{Channel Semantics Tokenizer.}
Raw channel names provide recording identity cues that can inform how physiological signals are modeled. To this end, we develop a channel semantics Tokenizer consists of byte-level embeddings, a convolutional stem, Transformer blocks, and padding-masked pooling followed by layer normalization. For channel
$c$ of modality $m$, its name $n^{(t)}_{m,c}$ is converted into a UTF-8 byte-token sequence $\textbf{b}^{(t)}_{m,c} = (b^{(t)}_{m,c,1}, ..., b^{(t)}_{m,c,L_{m,c}})$, where $L_{m,c}$ denotes the sequence length, each byte is mapped to a token in $\{1, ..., 256\}$, and $0$ is reserved for padding. This byte level representation preserves the original lexical structure, including case, whitespace, punctuation and byte order. 
The convolutional stem and Transformer blocks encode the byte sequence into contextual states $\textbf{H}^{(t)}_{m,c} \in \mathbb{R}^{L_{m,c} \times D_n}$. Then, the encoder $E_{name}$ provides the channel-name representation:
\begin{equation}
    \mathbf{h}^{(t)}_{m,c} = \operatorname{LN}\left(\operatorname{Mean}_{valid}\left(E_{name} \left(\textbf{b}^{(t)}_{m,c}\right)\right) \right) \in \mathbb{R}^{D_n}.
\end{equation}
To endow $\textbf{h}^{(t)}_{m,c}$ with physiological semantics, we pretrain the Channel Semantics Tokenizer using structured channel annotations. Specifically, auxiliary heads predict semantic type, derivation, anatomical location, and sensor information, while an alias-contrastive objective aligns different channel names sharing the same semantic signature. For ordered derivations, direction-aware hard negatives further distinguish reversed channel orders. After pretraining, the auxiliary heads are discarded, the tokenizer is frozen, and a lightweight adapter learns to incorporate the fixed semantic representations into waveform modeling:
\begin{equation}
    \textbf{s}^{(t)}_{m,c} = \operatorname{GELU}\left(\textbf{W}_{\operatorname{adp}}\operatorname{LN}\left({h}^{(t)}_{m,c}\right) + \textbf{b}_{\operatorname{adp}}\right) ,
\end{equation}
where $\textbf{s}^{(t)}_{m,c} \in \mathbb{R}^{D}$ is trainable semantic conditioning for intra-modality and channel aggregation.

\textbf{Intra-Modality Encoder.}
To model dependencies among channels within each modality, we jointly encode channel and temporal context under semantic conditioning. Each valid channel $\mathbf{X}^{(t)}_{m,c,:}$ is divided into $P$ non-overlapping patches $\{\mathbf{x}^{(t)}_{m,c,p}\}_{p=1}^{P}$, which are mapped by a shared convolutional tokenizer to physiological tokens
$\mathbf{v}^{(t)}_{m,c,p}=T_{\operatorname{ts}}(\mathbf{x}^{(t)}_{m,c,p})\in\mathbb{R}^{D}$.
The resulting tokens are then processed by Transformer blocks with semantic modulation and joint channel-temporal self-attention.
Channel semantics guide how these physiological tokens interact by conditioning their hidden representation before self-attention. The semantic representation  $\mathbf{s}^{(t)}_{m,c}$ is projected into feature-wise modulation parameters $\mathbf{\gamma}^{(t)}_{m,c}$ and $\mathbf{\beta}^{(t)}_{m,c}$ that are shared across all temporal patches of channel c:
\begin{align}
[\mathbf{\gamma}^{(t)}_{m,c}, \mathbf{\beta}^{(t)}_{m,c}]
&= \mathbf{W}_{\operatorname{sem}}\mathbf{s}^{(t)}_{m,c}
+ \mathbf{b}_{\operatorname{sem}},
\mathbf{\tilde{z}}^{(t)}_{m,c,p}
&= \left(1 + \mathbf{\gamma}^{(t)}_{m,c}\right) \odot
\operatorname{LN}\left(\mathbf{z}^{(t)}_{m,c,p}\right)
+ \mathbf{\beta}^{(t)}_{m,c},
\end{align}
where $\mathbf{z}^{(t)}_{m,c,p}$ denotes the hidden representation of patch $p$.
Self-attention is performed on the modulated representations.
Applying the shared encoder separately to each modality yields contextualized physiological tokens  $\{\mathbf{z}^{(t)}_{m,c,p}\}_{c,p}$.

To recover local waveform information from intra-modality context, we mask selected valid physiological tokens while preserving their channel semantics. A prediction head then reconstructs the masked token representations as $\mathbf{\hat{v}}{i} = R{\operatorname{intra}}(\mathbf{z}_i)$, with the objective
\begin{equation}
    \mathcal{L}_{\operatorname{intra}} = \frac{1}{|\Omega_{\operatorname{intra}}|}\sum_{i \in \Omega_{\operatorname{intra}}} [1-\operatorname{cos}(\mathbf{\hat{v}}_{i}, \operatorname{sg}(\mathbf{v}_{i}))], 
\end{equation}
where $\Omega_{\operatorname{intra}}$ denotes the set of valid masked positions, $\mathbf{v}_{i}$ is the tokenizer output before masking, and $\operatorname{sg}(\cdot)$ stops gradients through the reconstruction target.

After intra-modality encoding, a semantic-aware gate aggregates channel representations into modality-level temporal sequences. For each patch $p$, the gate score for channel $c$ combines its contextual token $\mathbf{z}^{(t)}_{m,c,p}$ with the projected channel-semantic representation:
\begin{equation}
    e^{(t)}_{m,c,p} = \mathbf{w}^{\top} \left( \operatorname{LN}\left( \mathbf{z}^{(t)}_{m,c,p}\right) + \mathbf{W}_s\mathbf{s}^{(t)}_{m,c} \right) .
\end{equation}

The scores are normalized over the valid channels using the channel validity mask $A^{(t)}$, and the contextualized representations are aggregated as %
\begin{equation}
    \alpha^{(t)}_{m,c,p} = \frac{A^{(t)}_{m,c}\operatorname{exp}(e^{(t)}_{m,c,p})}{\sum^C_{c^{\prime}=1}A^{(t)}_{m,c^{\prime}}\operatorname{exp}(e^{(t)}_{m,c^{\prime},p})}, \mathbf{f}^{(t)}_{m,p} = \sum_c\alpha^{(t)}_{m,c,p}\mathbf{z}^{(t)}_{m,c,p},
\end{equation}
where $\mathbf{f}^{(t)}_{m,p} \in \mathbb{R}^D$ integrates valid channels while retaining the modality and temporal indices.

\textbf{Inter-Modality Encoder.} To capture dependencies between physiological modalities, we design inter-modality interaction among the aggregated modality-level representation  $\mathbf{f}^{(t)}_{m,p}$ at each temporal patch. The Inter-Modality Encoder consists of learnable modality embeddings and a Transformer. For each modality $m$, a learnable modality embedding $\mathbf{e}_m \in \mathbb{R}^D$ is added to $\mathbf{f}^{(t)}_{m,p}$ to indicate its modality identity. Then, the tokens at $p$ are jointly encoded:
\begin{equation}
    \mathbf{G}^{(t)}_{p} = E_{\operatorname{inter}}\left(\left[\mathbf{{f}}^{(t)}_{1,p}+ \mathbf{e}_1,...,\mathbf{{f}}^{(t)}_{M,p} + \mathbf{e}_M \right]\right) \in \mathbb{R}^{M \times D}.
\end{equation}
The resulting $\mathbf{g}^{(t)}_{m,p} \in \mathbb{R}^D$ denotes the contextualized representation of modality $m$ at temporal position $p$, while missing modalities are excluded from attention.

During pretraining, we further leverage recovery from inter-modality context by masking selected valid modality–temporal representations. For each masked position $i$, a prediction head reconstructs the corresponding channel-aggregated representation as $\mathbf{\hat{f}}_i = R_{\operatorname{inter}}(\mathbf{g}_i)$, with the objective
\begin{equation}
    \mathcal{L}_{\operatorname{inter}} = \frac{1}{|\Omega_{\operatorname{inter}}|}\sum_{i \in \Omega_{\operatorname{inter}}} [1-\operatorname{cos}(\mathbf{\hat{f}}_{i}, \operatorname{sg}(\mathbf{f}_{i}))],
\end{equation}
where $\Omega_{\operatorname{inter}}$ denotes the set of valid masked modality-temporal positions.

To obtain window-level representations while preserving modality-specific information, a temporal aggregator independently summarizes each sequence ${\mathbf{g}^{(t)}{m,p}}{p=1}^{P}$. The temporal weights are:
\begin{equation}
    a^{(t)}_{m,p} = \operatorname{softmax}_p\left(\mathbf{w}^\top_{\operatorname{temp}} \operatorname{LN}\left(\mathbf{g}^{(t)}_{m,p}\right)\right),
\end{equation}
and obtain the corresponding window-level modality representation as $\mathbf{u}^{(t)}_m = \sum^P_{p=1}a^{(t)}_{m,p}\mathbf{g}^{(t)}_{m,p}$. 

\subsection{Time-Frequency and Morphology Encoding}
Raw PSG waveforms exhibit heterogeneous temporal characteristics, including multi-scale oscillations, localized spectral variations, and waveform morphology. To explicitly capture these complementary structures, we construct three image views for each PSG window: a CWT scalogram for scale-dependent oscillatory patterns, an STFT spectrogram for localized spectral content, and a morphology map for local time-domain waveform characteristics. CWT and STFT provide complementary time-frequency descriptions~\citep{moca2021time}, while the morphology view preserves waveform information beyond spectral transforms.

\textbf{Temporal Images Generation.} To convert multichannel PSG signals into consistently organized image inputs, the Image Creator consists of channel-wise feature transformations and modality-wise spatial composition.
For each valid channel $c$ of modality $m$, we construct a feature map
\begin{equation}
    \mathbf{Y}^{(t)}_{v,m,c} = \mathcal{T}_v\left(\mathbf{X}^{(t)}_{m,c,:}\right), v\in \{\operatorname{cwt}, \operatorname{stft}, \operatorname{morph}\},
\end{equation}
where $\mathcal{T}_v$ denotes the corresponding transformation. The CWT and STFT views apply a Morlet wavelet and a Hann analysis window, respectively, followed by the log-power mapping $\log(1 + |\cdot|^2)$.

To accommodate varying channel configurations while preserving a fixed modality layout, valid channels within each modality are aggregated using nonlearned weights derived from window-level signal statistics. CWT maps are normalized and resized before aggregation, whereas STFT and morphology maps are processed after aggregation. This yields a modality-specific image strip $\textbf{J}^{(t)}_{v,m} \in [0,1]^{H \times W}$, and the complete view is formed by vertically stacking the $M$ modality strips:
\begin{equation}
    I^{(t)}_v = \operatorname{Concat}_{\operatorname{height}}\left(\mathbf{J}^{(t)}_{v,1}, ..., \mathbf{J}^{(t)}_{v,M}\right) \in [0,1]^{(MH) \times W}.
\end{equation}

Missing modalities retain blank strips and are marked as invalid by spatial validity masks. All transformations are performed offline, and the image set $\mathcal{I}^{(t)} = \{I^{(t)}_{\operatorname{cwt}}, I^{(t)}_{\operatorname{stft}}, I^{(t)}_{\operatorname{morph}}\}$, together with its validity masks, serves as the input to the Multi-Image Encoder.

\textbf{Multi-Image Encoder.} 
To accommodate these differences while learning shared contextual features, the Multi-Image Encoder consists of view-specific convolutional stems and a parameter-shared Transformer. Each stem $\phi_v$ maps its input into spatial tokens $\mathbf{R}^{(t)}_v = \phi_v\left(I^{(t)}_v\right) \in \mathbb{R}^{Q \times D}$, where $Q$ is the number of spatial patches. A learnable view embedding $\mathbf{e}_v \in \mathbb{R}^D$ is added to preserve view identity, after which the tokens are passed through a Transformer $f_{\operatorname{img}}$ shared across the three views. We denote the resulting contextualized representation as $\mathbf{H}^{(t)}_v = f_{\operatorname{img}}(\mathbf{R}^{(t)}_v + \mathbf{e}_v) \in \mathbb{R}^{Q \times D}$. 

To preserve local structure, we apply masked latent reconstruction to the Multi-Image Encoder. Valid patches are masked before encoding and replaced with a learnable mask token, while a prediction head reconstructs their target latents. For position $q$ in view $v$, the target is $\mathbf{y}^{(t)}{v,q}=\mathbf{R}^{(t)}{v,q}+\mathbf{e}_v$, obtained from the unmasked image using the same view-specific stem under stop-gradient:
\begin{equation}
     \mathcal{L}_{\operatorname{mim}} = \frac{1}{|\Omega_{\operatorname{img}}|}\sum_{i \in \Omega_{\operatorname{img}}} [1-\operatorname{cos}(\mathbf{\hat{y}}_{v,q}, \operatorname{sg}(\mathbf{y}_{v,q}))],
\end{equation}
where $\Omega_{\operatorname{img}}$ contains the valid masked positions, $\mathbf{\hat{y}}_{v,q}$ denotes the prediction.

Local reconstruction captures view-specific spatial structure but does not explicitly align different transformations of the same PSG window. We therefore introduce cross-view contrastive learning at the representation level. The contextualized tokens are averaged over valid spatial positions:
\begin{equation}
    \mathbf{r}^{(t)}_v = \frac{1}{|\mathcal{Q}^{t}_v|}\sum_{q\in\mathcal{Q}^{t}_v}\mathbf{H}^{(t)}_{v,q} ,
\end{equation}
where $\mathcal{Q}^{t}_v$ denotes the valid spatial positions in view $v$. Representations from the same PSG window form positive pairs across views, while other eligible windows in the batch provide negatives. For a view pair $(v,v^{\prime})$, let $N$ denote the number of samples for which both views are valid. The directional contrastive loss and its symmetric aggregation are 
\begin{align}
     \mathcal{L}_{v\rightarrow v^{\prime}} &= -\frac{1}{N}\sum^{N}_{i=1}\operatorname{log}\frac{\operatorname{exp}(\operatorname{sim}(\mathbf{r}_{v,i},\mathbf{r}_{v^{\prime},i})/\tau)}{\sum^N_{j=1}\operatorname{exp}(\operatorname{sim}(\mathbf{r}_{v,i},\mathbf{r}_{v^{\prime},j})/\tau)},
     \mathcal{L}_{\operatorname{view}} &= \frac{1}{\mathcal{P}}\sum_{(v,v^{\prime}) \in \mathcal{P}}\frac{\mathcal{L}_{v\rightarrow v^{\prime}} + \mathcal{L}_{v^{\prime}\rightarrow v}}{2},
\end{align}
where $\mathbf{r}_{v,i}$ denotes the view-level representation of sample $i$, $\operatorname{sim}(\cdot,\cdot)$ is cosine similarity between $L_2$-normalized representations, $\tau$ is the temperature, and $\mathcal{P}=\{(\operatorname{cwt, stft}), (\operatorname{cwt, morph}), (\operatorname{stft, morph})\}$ contains the three complementary view pairs.

After cross-view representation learning, the view-level representations are integrated into a single image representation using learned sample-dependent weights:
\begin{equation}
    a^{(t)}_v = \operatorname{softmax}_{v\in\mathcal{V}^{(t)}}\mathbf{w}^\top_{\mathrm{img}}\operatorname{LN}(\mathbf{r}^{(t)}_v),~\mathbf{z}^{(t)}_{\mathrm{img}} = \sum_{v\in\mathcal{V}^{(t)}}\alpha^{(t)}_v\mathbf{r}^{(t)}_v \in \mathbb{R}^D,
\end{equation}
where $\mathcal V^{(t)}$ denotes the set of valid views and $\mathbf {w}_{\mathrm{img}}\in\mathbb{R}^D$ is a shared learnable scoring vector.

\subsection{Cross-Modal Alignment: Bridging Time-Series and Image Representations}

While multi-image views captures temporal dynamics, the semantic time-series view captures waveform dynamics and modality structure. To align these two views,
we average valid modality representations only along the alignment pathway and project the pooled time-series and aggregated image representations into a shared contrastive space:
\begin{equation}
    \mathbf{\bar{u}}^{(t)} = \frac{\sum^M_m a^{(t)}_m \mathbf{u}^{(t)}_m}{\sum^M_m a^{(t)}_m}, \mathbf{q}^{(t)}_{\operatorname{ts}} = g_{\operatorname{ts}}\left(\mathbf{\bar{u}}^{(t)}\right), \mathbf{q}^{(t)}_{\operatorname{img}} = g_{\operatorname{img}}\left(\mathbf{z}^{(t)}_{\operatorname{img}}\right),
\end{equation}
where $a^{(t)}_m$ indicates the validity of modality $m$, and $g_{\operatorname{ts}}$ and $g_{\operatorname{img}}$ are modality projection heads. 

For $N$ paired windows, same-windows form positives and others form negatives. We use cosine similarity between $L_2$-normalized projections to define the cross-modal contrastive objective:
\begin{align}
     \mathcal{L}_{\mathrm{ts}\rightarrow \mathrm{img}} &= -\frac{1}{N}\sum^{N}_{i=1}\log\frac{\operatorname{exp}(\operatorname{sim}(\mathbf{q}_{\mathrm{ts},i},\mathbf{q}_{\mathrm{img},i})/\tau)}{\sum^N_{j=1}\operatorname{exp}(\operatorname{sim}(\mathbf{q}_{\mathrm{ts},i},\mathbf{q}_{\mathrm{img},j})/\tau)},
     \mathcal{L}_{\mathrm{align}} &= \frac{\mathcal{L}_{\mathrm{ts}\rightarrow \mathrm{img}} + \mathcal{L}_{\mathrm{img}\rightarrow \mathrm{ts}}}{2}.
\end{align}
The final representation is
$Z^{(t)} = [\mathbf{u}^{(t)}_1, \ldots, \mathbf{u}^{(t)}_M, \mathbf{z}^{(t)}_{\mathrm{img}}] \in \mathbb{R}^{(M+1)\times D}$,
formed by stacking the unprojected time-series and image representations. Combining cross-modal alignment with the reconstruction and cross-view objectives, the overall pretraining loss is:
\begin{equation}
    \mathcal{L}
    = \mathcal{L}_{\mathrm{intra}}
    + \lambda_{\mathrm{inter}}\mathcal{L}_{\mathrm{inter}}
    + \lambda_{\mathrm{mim}}\mathcal{L}_{\mathrm{mim}}
    + \mathcal{L}_{\mathrm{view}}
    + \mathcal{L}_{\mathrm{align}},
\end{equation}
where $\lambda_{\mathrm{inter}}$ is progressively increased during pretraining to strengthen cross-modal reconstruction, $\lambda_{\mathrm{mim}}$ controls the contribution of masked image reconstruction.

\section{Experiments}
\label{sec：exp}
We evaluate six downstream tasks, with the four task results in Sections~\ref{sec:st}--~\ref{sec:msv}, two tasks results are reported in Appendix~\ref{a:ahi}--~\ref{a:de}, and additional results are provided in Appendix~\ref{a:ss}--~\ref{a:apr}.

\subsection{Dataset Usage}

\begin{wrapfigure}{r}{0.45\linewidth}
\centering
\vspace{-1cm}
\includegraphics[width=\linewidth]{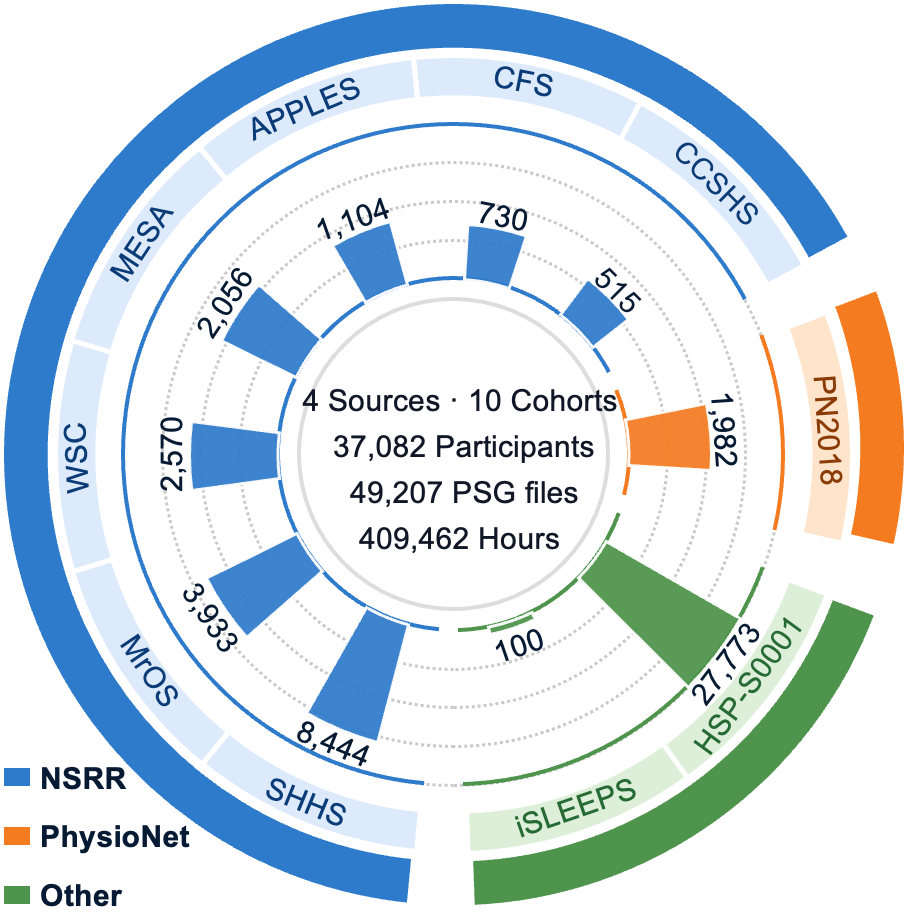}
\vspace{-0.5cm}
\caption{Statistics of the PSG datasets.}
\label{fig:data}
\vspace{-0.3cm}
\end{wrapfigure}

As shown in Figure~\ref{fig:data}, we construct a heterogeneous PSG corpus from 10 cohorts, comprising 37,082 participants, 49,207 recordings, and 409,462 hours of sleep recordings.
\textit{Pretraining:}
SemPSG is pretrained on nine cohorts: SHHS~\citep{quan1997shhs}, MrOS~\citep{blackwell2011mros}, WSC~\citep{young2009wsc}, MESA~\citep{chen2015mesa}, APPLES~\citep{quan2011apples}, CCSHS~\citep{rosen2003ccshs}, PN2018~\citep{ghassemi2018cinc}, HSP-S0001~\citep{li2026hsp}, and iSLEEPS~\citep{maiti2026isleep}. \textit{For cohorts also used downstream, only training subjects are used for pretraining to prevent leakage.}
\textit{Downstream Evaluation:}
We evaluate on SHHS, APPLES, and CFS. SHHS and APPLES assess within-cohort transfer, while CFS~\citep{redline1995cfs} is excluded from pretraining and serves as an external cohort for cross-dataset generalization. For SHHS and CFS, we follow the partition protocol of SleepMaMi~\citep{park2026sleepmami}.

\subsection{Sleep Staging}
\label{sec:st}
Sleep staging is formulated as a five-class classification task over 30-s PSG epochs. We evaluate both in-cohort generalization on SHHS1 and cross-dataset generalization on the unseen CFS cohort. As shown in Table~\ref{tab:sleep_staging}, SemPSG consistently outperforms other sleep foundation models across both datasets and all metrics, with the largest gain on SHHS1, improving Kappa by 5.2 percentage points over SleepMaMi. General-purpose time-series models remain competitive, with Zeus performing particularly well on CFS, indicating that generic temporal representations capture useful staging information. SemPSG further improves performance by explicitly modeling intra-modality channel dependencies and inter-modality physiological interactions.

\subsection{Sleep-Disordered Breathing Analysis}

We evaluate sleep-disordered breathing (SDB) using two downstream tasks, SDB classification and apnea–hypopnea index (AHI) regression. SDB classification identifies normal and disordered breathing on a second-by-second basis, whereas AHI regression estimates overall respiratory disturbance severity at the recording level. As shown in Tables~\ref{tab:sleep_disorder_classification_shhs1}, %
SemPSG performs strongly on the SDB classification, which improves Macro-F1 by 9.6 percentage points over the second-best model SleepMaMi. For the experimental results of AHI regression, please see Appendix~\ref{a:ahi}

\begin{table*}[t]
\centering
\caption{Sleep staging performance on SHHS1 and CFS. All metrics are reported as percentages.}
\label{tab:sleep_staging}
\renewcommand{\arraystretch}{0.8}
\begin{tabular*}{0.98\textwidth}{
@{\extracolsep{\fill}}
c c l c c c
@{}
}
\toprule
\textbf{Dataset} & \textbf{Category} & \textbf{Model}
& \textbf{Accuracy $\uparrow$} & \textbf{Kappa $\uparrow$} & \textbf{Macro-F1 $\uparrow$} \\
\midrule

\multirow{7}{*}{SHHS1}
& \multirow{3}{*}{\shortstack{General-purpose TSFM}}
& MOMENT & 79.4 & 70.0 & 65.6 \\
& & UniTS & 64.2 & 53.3 & 59.2 \\
& & Zeus & 79.2 & \underline{70.5} & 69.5 \\
\cmidrule(lr){2-6}

& EEG Foundation Model 
& LaBraM & 70.2 & 55.7 & 54.1 \\
\cmidrule(lr){2-6}

& \multirow{3}{*}{\shortstack{Sleep Foundation Model}}
& SleepFM  & 69.7 & 55.9 & 56.3 \\
& & SleepMaMi & \underline{81.9} & 70.0 & \underline{74.1} \\
& & SemPSG & \textbf{82.4} & \textbf{75.2} & \textbf{74.6} \\

\midrule

\multirow{7}{*}{CFS}
& \multirow{3}{*}{\shortstack{General-purpose TSFM}}
& MOMENT & 71.4 & 56.7 & 52.4 \\
& & UniTS & 64.1 & 53.3 & 58.1 \\
& & Zeus & \underline{80.8} & \underline{72.9} & 69.7 \\
\cmidrule(lr){2-6}

& EEG Foundation Model
& LaBraM & 77.5 & 67.3 & 61.8 \\
\cmidrule(lr){2-6}

& \multirow{3}{*}{\shortstack{Sleep Foundation Model}}
& SleepFM & 71.1 & 57.5 & 57.4 \\
& & SleepMaMi & 80.7 & \underline{72.9} & \underline{71.2} \\
& & SemPSG & \textbf{81.9} & \textbf{74.8} & \textbf{73.6} \\

\bottomrule
\vspace{-0.5cm}
\end{tabular*}
\end{table*}

\begin{table*}[t]
\centering
\caption{Sleep-disordered breathing classification performance on SHHS1 and CFS.%
}
\label{tab:sleep_disorder_classification_shhs1}
\renewcommand{\arraystretch}{0.8}
\begin{tabular*}{0.98\textwidth}{
@{\extracolsep{\fill}}
c c l c c
@{}
}
\toprule
\textbf{Dataset} & \textbf{Category} & \textbf{Model}
& \textbf{Accuracy $\uparrow$} & \textbf{Macro-F1 $\uparrow$} \\
\midrule

\multirow{6}{*}{SHHS1}
& \multirow{2}{*}{General-purpose TSFM}
& MOMENT & 73.4 & 33.4 \\
& & UniTS & \textbf{88.2} & 48.8 \\
\cmidrule(lr){2-5}

& EEG Foundation Model
& LaBraM & 50.2 & 20.4 \\
\cmidrule(lr){2-5}

& \multirow{3}{*}{Sleep Foundation Model}
& SleepFM & 77.5 & 39.4 \\
& & SleepMaMi & 77.3 & \underline{60.6} \\
& & SemPSG & \underline{83.5} & \textbf{70.2} \\

\midrule

\multirow{6}{*}{CFS}
& \multirow{2}{*}{General-purpose TSFM}
& MOMENT & 74.2 & 38.5 \\
& & UniTS & \underline{85.5} & 40.0 \\
\cmidrule(lr){2-5}

& EEG Foundation Model
& LaBraM & 47.0 & 41.7 \\
\cmidrule(lr){2-5}

& \multirow{3}{*}{Sleep Foundation Model}
& SleepFM & 82.1 & 46.6 \\
& & SleepMaMi & 79.9 & \underline{66.1} \\
& & SemPSG & \textbf{86.5} & \textbf{69.4} \\

\bottomrule
\end{tabular*}
\end{table*}

\begin{table*}[t]
\centering
\caption{Disease prediction performance measured on SHHS1.}
\label{tab:disease_prediction}
\setlength{\tabcolsep}{5pt}
\renewcommand{\arraystretch}{0.8}
\begin{tabular}{llcccccc}
\toprule
\multirow{3}{*}{\textbf{Category}}
& \multirow{3}{*}{\textbf{Models}}
& \multicolumn{6}{c}{\textbf{Disease Outcomes}} \\
\cmidrule(lr){3-8}

& & \textbf{Angina} & \textbf{CVD Death}
& \textbf{CHF} & \textbf{CHD Death}
& \textbf{MI} & \textbf{Stroke} \\
\cmidrule(lr){3-8}

& & \multicolumn{6}{c}{\textit{C-Index $\uparrow$}} \\
\midrule

\multirow{2}{*}{\shortstack[l]{General-purpose\\TSFM}}
& MOMENT & 0.676 & 0.698 & 0.692 & 0.689 & 0.584 & 0.609 \\
& UniTS & 0.593 & 0.640 & 0.636 & 0.667 & 0.579 & 0.620 \\
\midrule

\multirow{3}{*}{\shortstack[l]{Sleep Foundation\\Model}}
& SleepFM
& 0.632 & \underline{0.791} & 0.764
& \underline{0.781} & 0.636 & \underline{0.729} \\

& SleepMaMi
& \textbf{0.778} & 0.788 & \underline{0.793}
& 0.776 & \textbf{0.662} & 0.718 \\

& SemPSG
& \underline{0.727} & \textbf{0.836} & \textbf{0.798}
& \textbf{0.837} & \underline{0.660} & \textbf{0.751} \\

\bottomrule
\vspace{-0.5cm}
\end{tabular}
\end{table*}

\begin{table*}[t]
\centering
\caption{Cognition and affective assessment performance on APPLES.}
\label{tab:cognition_emotion}
\renewcommand{\arraystretch}{0.8}
\begin{tabular*}{0.98\textwidth}{
@{\extracolsep{\fill}}
c c l c c
@{}
}
\toprule
\textbf{Dataset} & \textbf{Category} & \textbf{Model}
& \textbf{HAM-D MAE $\downarrow$}
& \textbf{WASI MAE $\downarrow$} \\
\midrule

\multirow{5}{*}{APPLES}
& \multirow{2}{*}{General-purpose TSFM}
& MOMENT & 3.15 & 9.12 \\
& & UniTS & 3.13 & 8.86 \\
\cmidrule(lr){2-5}

& \multirow{3}{*}{Sleep Foundation Model}
& SleepFM & \underline{3.02} & \underline{8.56} \\
& & SleepMaMi & 3.05 & 9.88 \\
& & SemPSG & \textbf{2.84} & \textbf{8.00} \\

\bottomrule
\end{tabular*}
\end{table*}

\subsection{Disease Prediction}
    We evaluate whether the learned PSG representations transfer to downstream disease prediction. Based on the outcomes available in SHHS1, we consider six cardiovascular and cerebrovascular endpoints: angina, CVD death, CHF, CHD death, MI, and stroke. Performance is evaluated using the concordance index (C-index). As shown in Table~\ref{tab:disease_prediction}, SemPSG achieves the highest C-index for four of the six outcomes, including CVD death, CHF, CHD death, and stroke. The largest gains are observed for CHD death and CVD death, where SemPSG outperforms the second-best model, SleepFM, by 7.2\% and 5.7\%, respectively. Its leading performance on CHF and stroke further indicates that the gains extend across multiple cardiovascular and cerebrovascular endpoints. 

\subsection{Cognition and Affective Assessment}
We assess cognitive and affective information in the learned PSG representations via HAM-D
and WASI
score prediction on APPLES. As shown in Table~\ref{tab:cognition_emotion}, SemPSG achieves the lowest MAE on both tasks, reducing MAE by 6.0\% for HAM-D and 6.5\% for WASI over the second-best SleepFM. These results demonstrate that SemPSG preserves information relevant to both affective symptoms and cognitive function beyond conventional sleep outcomes.

\subsection{Modality Semantics Visualization}
\label{sec:msv}

\begin{figure}[!t]
    \centering
    \includegraphics[width=0.95\linewidth]{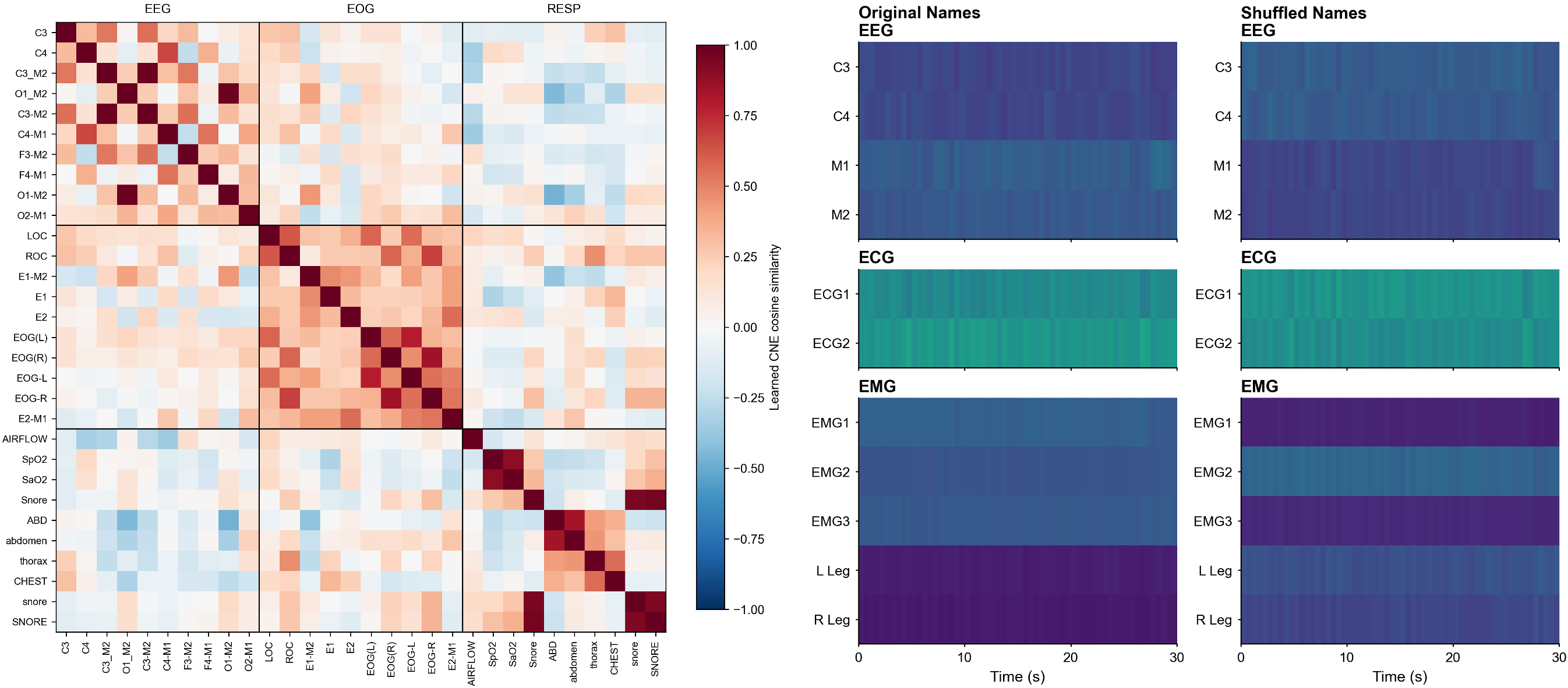}
    \vspace{-0.5cm}
    \caption{Channel and modality semantics visualization.
    Left: Pairwise cosine similarity between learned channel-semantic embeddings. Right: Learned channel-gating weights over a 30-s window for EEG, ECG, and EMG under the original and shuffled channel-name assignments.
    }
    \label{fig:heatmap}
\end{figure}

\begingroup
\setlength{\intextsep}{4pt}
\setlength{\abovecaptionskip}{4pt}
\begin{wrapfigure}{r}{0.4\linewidth}
    \centering
    \includegraphics[width=\linewidth]{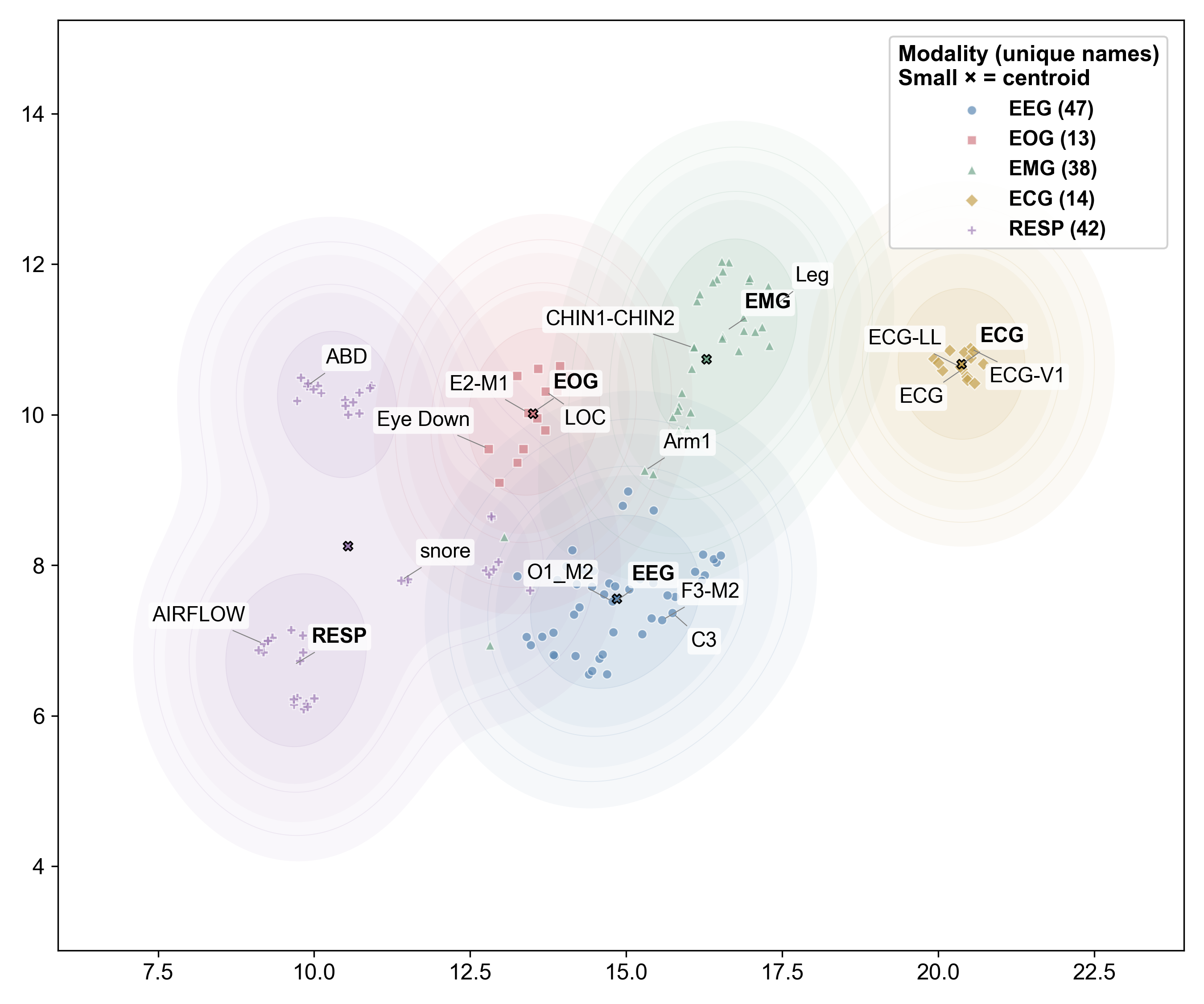}
    \caption{UMAP of channel semantic embeddings across modalities.}
    \label{fig:channel_semantic_umap}
\end{wrapfigure}

We visualize the representation of channel semantics tokenizer to explain the semantics of PSG.
We assess the effect of channel semantics on channel fusion by shuffling channel-name assignments while keeping waveforms unchanged. As shown in Figure~\ref{fig:heatmap} (right), the gating patterns change for EEG, ECG, and EMG, demonstrating that channel-name semantics directly influence intra-modality channel weighting.
Figure~\ref{fig:heatmap} (left) shows that channels within the same physiological modality generally exhibit higher semantic similarity.
Figure~\ref{fig:channel_semantic_umap} reveals finer-grained intra-modality structure. 
For example, respiratory channels form distinct subregions rather than collapsing into a single cluster, indicating that the learned space captures both modality-level and channel-specific semantics.
\par
\endgroup

\section{Conclusion}

In this work, we present \textbf{SemPSG}, a semantic-aware foundation model for learning generalizable representations from heterogeneous PSG recordings. SemPSG integrates channel semantics, captures both intra- and inter-modality dependencies, and aligns waveform and image representations in a shared representation space. Experiments across a broad range of downstream tasks show that the learned representations transfer effectively across different prediction settings, supporting the use of SemPSG as a general-purpose foundation model for sleep and physiological signal analysis.

\subsection*{AI use statement}
Generative AI tools were used to assist with language polishing and with limited implementation and code-related tasks, including code editing and debugging. The research methodology, model architecture, experimental design, and interpretation of results were developed by the authors.

\subsection*{Ethics statement}
This study uses de-identified PSG recordings from publicly available datasets and involves no new human-subject data collection. Ethical approval and informed consent for the original studies were obtained by the respective institutions, and all datasets were used in accordance with their access and usage requirements.


\bibliography{reference}
\bibliographystyle{iclr2027_conference}

\appendix
\section{Appendix}

\subsection{Related Work}

\textbf{Time-Series Representation Learning} aims to extract generalizable features from sequential data that can be transferred across downstream tasks. To reduce the dependence on task-specific annotations, self-supervised learning has become a widely adopted paradigm, with representative approaches based on temporal-view consistency~\citep{yue2022ts2vec,ma2026phat} and masked reconstruction~\citep{dong2023simmtm, cheng2026timemae,zhao2025stem}. More recently, general-purpose time series foundation models have further scaled representation learning to heterogeneous datasets and tasks. For example, MOMENT~\citep{goswami2024moment} learns generalizable representations through large-scale masked pretraining, while UniTS~\citep{gao2024units} adopts a unified architecture across different time series tasks. These models demonstrate the potential of large-scale pretraining in general-purpose time series analysis, but they are not specifically designed to capture the structured physiological information associated with PSG channels.

\textbf{Foundation Models for Polysomnography Analysis} have recently emerged to exploit the heterogeneous physiological signals and temporally structured sleep patterns in PSG. For instance, SleepFM~\citep{thapa2026multimodal} learns representations through contrastive learning across physiological signals, while SleepMami~\citep{park2026sleepmami} models sleep patterns at multiple temporal scales. Other studies have explored representation learning across heterogeneous nocturnal biosignals~\citep{yuan2026sleep2vec} and channel-adaptive time-frequency modeling~\citep{huang2026unified}, further broadening the range of PSG signal settings considered. Despite these advances, existing approaches do not explicitly represent PSG channel identity through its structured physiological attributes, such as signal type, recording location, and reference configuration. This limits their ability to exploit semantics among channels and transfer knowledge across heterogeneous channel configurations.

\subsection{Preliminaries}
\textbf{PSG Signals}. A PSG (Polysomnography) signal $\mathcal{R}$ contains synchronized physiological time-series signals from five modalities $\mathcal{M} = \{ EEG, EOG, EMG, ECG, RESP \}$. We divide each recording into non-overlapping 30s windows $\{\mathcal{W}^{(t)} \}^T_{t=1}$. For the $t$-th window, the time series signals are represented as $\mathbf{X}^{(t)} \in \mathbb{R}^{M \times C \times S}$, where $M = 5$, $C$ is the padded channel dimension, and $S$ is the number of samples per channel. The corresponding raw channel names are denoted by $\mathcal{N}^{(t)} = \{n^{(t)}_{m,c}\}$, with the channel validity indicated by $A^{(t)} \in \{ 0,1\}^{M \times C}$. Each window is also associated with $V = 3$ image views $\mathcal{I}^{(t)} = \{ \mathbf{I}^{(t)}_v\}^V_{v=1}$ derived from the same physiological signals. Thus, each PSG window is represented as $\mathcal{W}^{(t)} = (\mathbf{X}^{(t)}, \mathcal{N}^{(t)}, \mathcal{I}^{(t)}, A^{(t)})$.

\textbf{Problem Formulation}. Given a PSG window $\mathcal{W}^{t}$, our goal is to learn a foundation model $f_\theta$ that maps its time series signals, channel name, and image views to a unified representation, $Z^{(t)} = f_\theta(\mathcal{W}^{(t)}) = [\mathbf{u}^{(t)}_1, ..., \mathbf{u}^{(t)}_M, \mathbf{z}^{(t)}_{img}]$, where $Z^{(t)} \in \mathbb{R}^{(M + 1) \times D}$, $\mathbf{u}^{(t)}_m \in \mathbb{R}^{D}$ denotes the representation of physiological modality $m$, and $\mathbf{z}^{(t)}_{img} \in \mathbb{R}^{D}$ denotes the image representation. The foundation model is pretrained on unlabeled PSG windows with self-supervised objectives that capture local physiological structure and promote consistency across time-series and image representations.

\subsection{Physiological Modalities and Channel Organization in PSG}

PSG provides a multimodal view of sleep by jointly recording neurophysiological and cardiorespiratory activity. We organize the recorded signals into five physiological modalities: electroencephalography (EEG), electrooculography (EOG), electromyography (EMG), electrocardiography (ECG), and respiration (RESP).

\textbf{Neurophysiological Signals.}
EEG, EOG, and EMG jointly characterize brain activity, eye movements, and muscle tone during sleep, providing complementary information about neurophysiological states.

\textbf{Cardiorespiratory Signals.}
ECG captures cardiac activity and autonomic variations, while RESP may include airflow, thoracic and abdominal effort, and SpO$_2$, jointly characterizing respiratory dynamics during sleep.

\textbf{Channel Organization.}
Each modality may contain multiple channels with different recording locations, derivations, and sensor characteristics. We retain these channel-level distinctions within a consistent modality space, motivating the modeling of both intra-modality channel dependencies and inter-modality physiological interactions.

\begin{figure}[!h]
    \centering
    \includegraphics[width=0.75\linewidth]{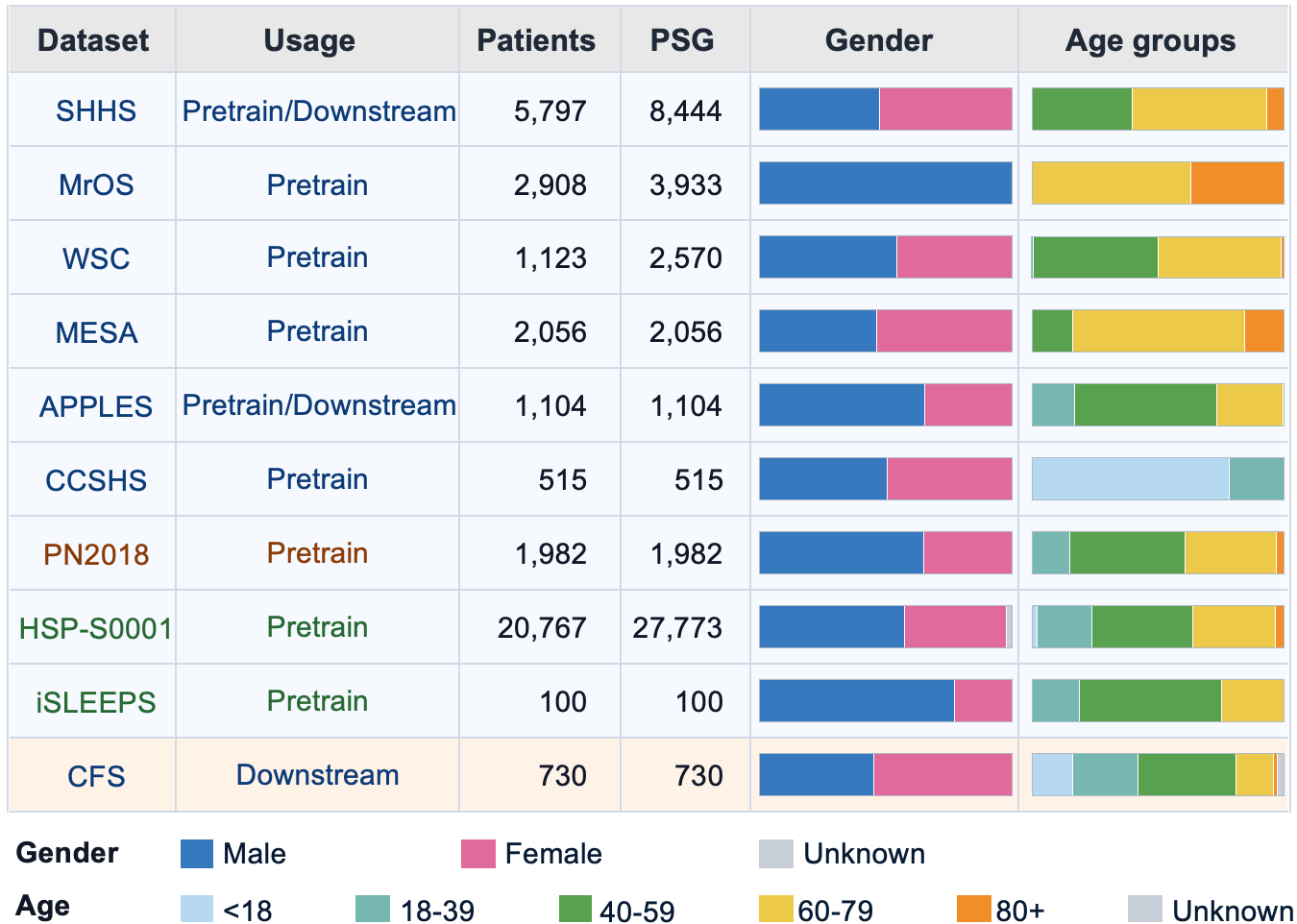}
\caption{\textbf{Statistics and demographic characteristics of the PSG datasets used in this study.}
For each cohort, we report its usage in pretraining and downstream evaluation, the number of patients and PSG recordings, and the distributions of gender and age groups. SHHS and APPLES are used for both pretraining and downstream evaluation, CFS is reserved exclusively for downstream evaluation, and the remaining cohorts are used for pretraining. The colored bars visualize the demographic composition of each cohort.}
\label{fig:dataset-tab}
\end{figure}

\subsection{Dataset Details}
We pretrain SemPSG on multiple PSG cohorts collected from diverse populations and recording settings to expose the model to substantial variation in physiological signals, participant characteristics, and channel configurations during representation learning. As summarized in Figure~\ref{fig:dataset-tab}, the pretraining corpus includes SHHS~\citep{quan1997shhs}, MrOS~\citep{blackwell2011mros}, WSC~\citep{young2009wsc}, MESA~\citep{chen2015mesa}, APPLES~\citep{quan2011apples}, CCSHS~\citep{rosen2003ccshs}, PN2018~\citep{ghassemi2018cinc}, HSP-S0001~\citep{li2026hsp}, and iSLEEPS~\citep{maiti2026isleep}. These cohorts differ markedly in scale, demographic composition, clinical characteristics, acquisition environments, and PSG channel configurations
SHHS and APPLES contribute to both pretraining and downstream evaluation, with evaluation subjects excluded from representation learning to prevent information leakage. The remaining cohorts are used exclusively for pretraining, further broadening the diversity of the training distribution. For downstream evaluation, we additionally include CFS~\citep{redline1995cfs}, which is completely excluded from pretraining and therefore serves as an unseen cohort for assessing cross-dataset generalization. Below, we describe the characteristics of each cohort and its specific role in our experimental setup.

\begin{itemize}
    \item \textbf{Sleep Heart Health Study (SHHS)}. SHHS is a large-scale cohort study containing overnight PSG recordings together with extensive demographic and clinical information. The recordings provide multiple physiological signals covering neurophysiological, cardiac, and respiratory activity. In this study, SHHS is included in the pretraining corpus and is also used for downstream evaluation. Subjects assigned to downstream test sets are kept separate from the recordings used for representation learning.
    
    \item \textbf{MrOS Sleep Study (MrOS)}. The MrOS Sleep Study is an ancillary study of the Osteoporotic Fractures in Men cohort and contains overnight PSG recordings from community-dwelling older men. The sleep assessment includes unattended home PSG with neurophysiological, cardiac, and respiratory measurements, providing recordings from an older population and a home-based acquisition setting. We include MrOS as part of the pretraining corpus.
    
    \item \textbf{Multi-Ethnic Study of Atherosclerosis (MESA)}. MESA includes overnight sleep recordings collected as part of a population-based study of cardiovascular health. Its PSG recordings contain signals from multiple physiological modalities and provide additional variation in participant characteristics and recording configurations. We include MESA as part of the pretraining corpus.

    \item \textbf{Wisconsin Sleep Cohort (WSC)}. WSC is a longitudinal observational study established to investigate the prevalence, causes, and consequences of sleep disorders in adults. It includes repeated overnight in-laboratory PSG recordings collected from community-dwelling participants, together with extensive demographic and clinical information. The cohort therefore contributes longitudinal PSG data and additional variation in participant characteristics and recording sessions. We include WSC as part of the pretraining corpus.
    
    \item \textbf{Apnea Positive Pressure Long-term Efficacy Study (APPLES)}. APPLES is a multicenter clinical study of obstructive sleep apnea originally conducted to investigate the long-term effects of continuous positive airway pressure therapy. The dataset contains diagnostic PSG recordings together with demographic, clinical, and neurocognitive information from participants with suspected or diagnosed obstructive sleep apnea. In this study, APPLES is included in the pretraining corpus and is also used for downstream evaluation. Subjects assigned to downstream test sets are excluded from pretraining to maintain separation between representation learning and evaluation.
    
    \item \textbf{Cleveland Children's Sleep and Health Study (CCSHS)}. CCSHS is a population-based cohort established to investigate sleep and sleep-disordered breathing in children and adolescents. The available TREC visit includes full overnight laboratory PSG recordings with multiple physiological signals, introducing a younger population and a laboratory-based recording setting into the pretraining data. We include CCSHS as part of the pretraining corpus.

    \item \textbf{PhysioNet/Computing in Cardiology Challenge 2018 (PN2018)}. PN2018 contains overnight PSG recordings collected from subjects monitored in the Massachusetts General Hospital sleep laboratory for the diagnosis of sleep disorders. The recordings include multiple physiological signals together with expert annotations of sleep stages and sleep-related arousals. We use the physiological recordings for self-supervised pretraining, without relying on the accompanying task annotations.

    \item \textbf{Human Sleep Project (HSP). }The Human Sleep Project is a large-scale clinical sleep database containing overnight PSG recordings collected from patients undergoing sleep assessment. In this study, we use the HSP-S0001 subset, which provides multichannel physiological recordings from a clinical sleep population and further expands the diversity of recording conditions and channel configurations represented in the pretraining corpus. HSP-S0001 is used exclusively for pretraining.

    \item \textbf{iSLEEPS.} iSLEEPS is a clinical PSG dataset collected from patients with ischemic stroke at the National Institute of Mental Health and Neurosciences in India. It contains full-night PSG recordings acquired in a clinical setting, together with expert sleep-stage annotations, respiratory events, and associated demographic and clinical information. The recordings capture multiple physiological signals across neurophysiological, cardiac, and respiratory modalities, providing PSG data from a population with underlying neurological conditions. Compared with population-based sleep cohorts, iSLEEPS introduces additional variation in participant characteristics, disease status, and recording conditions. 
    We include iSLEEPS exclusively in the pretraining corpus to broaden the physiological and clinical diversity encountered during representation learning and to expose SemPSG to heterogeneous PSG patterns.

    \item \textbf{Cleveland Family Study (CFS).} CFS is a family-based cohort established to investigate the familial aggregation of sleep apnea and related physiological characteristics. The study includes participants from a broad age range, with full overnight PSG available from the later study visit. In our experiments, CFS is used exclusively for downstream evaluation and is not included in the pretraining corpus, ensuring that its recordings remain unseen during representation learning.

\end{itemize}

\par
\noindent
\begin{minipage}{\linewidth}

\subsection{Algorithms}
\label{sec:algorithms}

\small
\setlength{\intextsep}{6pt}
\algrenewcommand{\algorithmicindent}{1.2em}

\begin{algorithm}[H]
\caption{Channel semantics tokenizer Pretraining}
\label{alg:channel-semantics}
\begin{algorithmic}[1]
\Procedure{TrainSemantics}{$\mathcal{D}_{\mathrm{name}},\mathcal{R}$}
    \State Initialize encoder $T_{\phi}$ and semantic heads $C_{\psi}$
    \For{each batch $(N,Y,A)$}
        \State $h \gets T_{\phi}(N)$;
            \quad $z \gets \operatorname{Normalize}(h)$
        \State $\mathcal{L}_{\mathrm{sem}} \gets
            \Call{SemanticLoss}{C_{\psi}(h),Y,A}$
            \Comment{Attribute supervision}
        \State $\mathcal{L}_{\mathrm{alias}} \gets
            \Call{AliasContrast}{z,\mathcal{R}}$
            \Comment{Alias consistency}
        \State $\mathcal{L}_{\mathrm{dir}} \gets
            \Call{DirectionLoss}{T_{\phi},N,\mathcal{R}}$
            \Comment{Ordered derivations}
        \State $\mathcal{L} \gets
            \mathcal{L}_{\mathrm{sem}}
            + \mathcal{L}_{\mathrm{alias}}
            + \lambda_{\mathrm{dir}}\mathcal{L}_{\mathrm{dir}}$
        \State Update $(\phi,\psi)$ to minimize $\mathcal{L}$
    \EndFor
    \State \Return validation-selected $T_{\phi}$
\EndProcedure
\end{algorithmic}
\end{algorithm}

\begin{algorithm}[H]
\caption{SemPSG Pretraining Overview}
\label{alg:psgfm-pretraining}
\begin{algorithmic}[1]
\Procedure{PretrainPSGFM}{$\mathcal{D},T_{\phi}$}
    \State Freeze $T_{\phi}$; initialize trainable parameters $\theta$
    \State Prepare CWT, STFT, and morphology views
    \For{each batch $(X,N,I,A)$ at step $s$}
        \State $(U,\mathcal{L}_{\mathrm{intra}},
            \mathcal{L}_{\mathrm{inter}})
            \gets \Call{EncodeSignals}{X,N,A}$
        \State $(R,z_{\mathrm{img}},\mathcal{L}_{\mathrm{mim}})
            \gets \Call{EncodeImages}{I,A}$
        \State $\mathcal{L}_{\mathrm{view}} \gets
            \Call{ViewContrast}{R}$
            \Comment{Across image views}
        \State $\mathcal{L}_{\mathrm{align}} \gets
            \Call{Align}{U,z_{\mathrm{img}},A}$
            \Comment{Signals and images}
        \State $\mathcal{L} \gets
            \mathcal{L}_{\mathrm{intra}}
            + \lambda_{\mathrm{inter}}(s)\mathcal{L}_{\mathrm{inter}}
            + \lambda_{\mathrm{mim}}\mathcal{L}_{\mathrm{mim}}
            + \mathcal{L}_{\mathrm{view}}
            + \mathcal{L}_{\mathrm{align}}$
        \State Update $\theta$ to minimize $\mathcal{L}$
    \EndFor
    \State \Return $f_{\theta}$
\EndProcedure

\Statex
\Function{EncodeSignals}{$X,N,A$}
    \State $E \gets \operatorname{Adapter}(T_{\phi}(N))$;
        \quad $V \gets \Call{Tokenize}{X}$
    \State $(H,\mathcal{L}_{\mathrm{intra}})
        \gets \Call{MaskedIntra}{V,E,A}$
        \Comment{Within modalities}
    \State $F \gets \Call{SemanticPool}{H,E,A}$
        \Comment{Across channels}
    \State $(G,\mathcal{L}_{\mathrm{inter}})
        \gets \Call{MaskedInter}{F,A}$
        \Comment{Across modalities}
    \State $U \gets \Call{TemporalPool}{G,A}$
    \State \Return $(U,\mathcal{L}_{\mathrm{intra}},
        \mathcal{L}_{\mathrm{inter}})$
\EndFunction
\end{algorithmic}
\end{algorithm}

\end{minipage}
\par

\begin{figure}[h]
    \centering
    \includegraphics[width=1.0\textwidth]{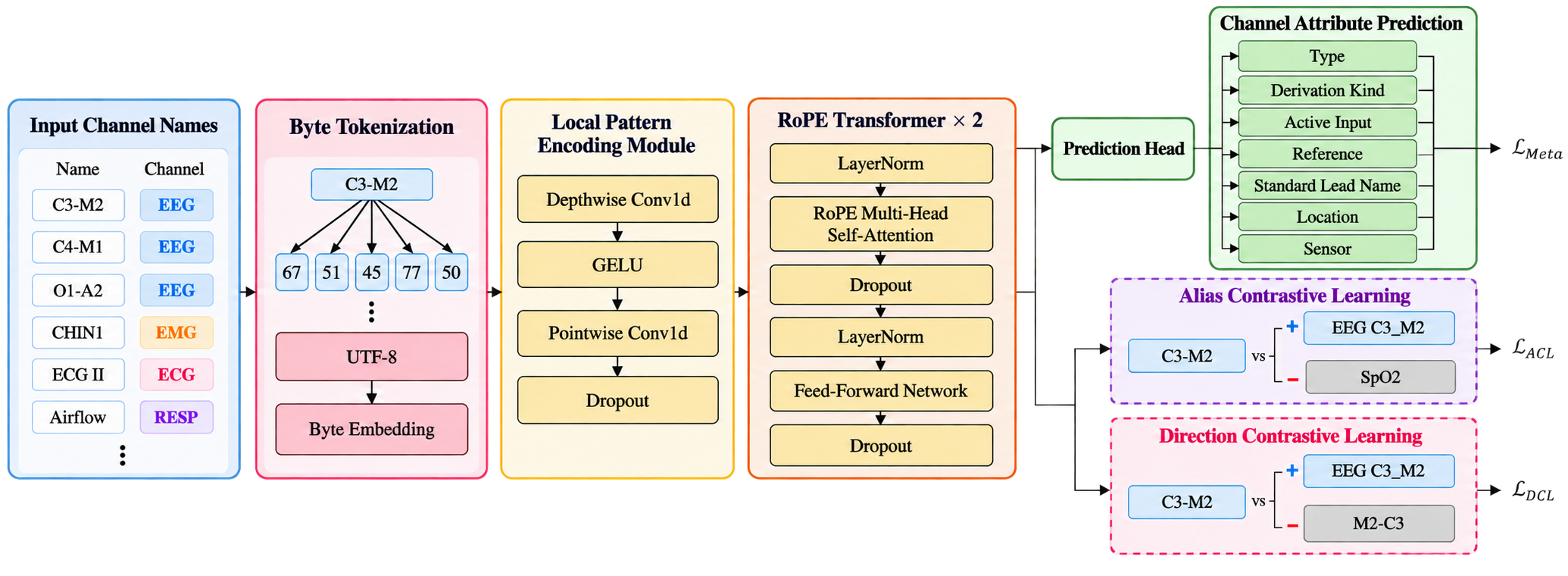}
    \caption{Channel semantics tokenizer training process.}
    \label{fig:cse}
\end{figure}

\subsection{Baselines}
We compare SemPSG with representative pretrained models from three categories, including general-purpose time-series foundation models, sleep-specific foundation models, and EEG foundation models where applicable. The general-purpose baselines include UniTS \citep{gao2024units}, MOMENT \citep{goswami2024moment}, and Zeus~\citep{fu2026zeus}, while the sleep-specific baselines include SleepFM \citep{thapa2026multimodal} and SleepMaMi \citep{park2026sleepmami}. LaBraM~\citep{jiang2024large} is additionally included as an EEG foundation model for tasks where it is applicable. Since not all pretrained models support every downstream setting, the exact set of baselines varies across tasks. All methods are evaluated using the same downstream data splits and task-specific evaluation protocols.

\begin{itemize}
    \item \textbf{UniTS}. UniTS is a general-purpose time-series model designed to handle multiple predictive and generative tasks within a unified architecture. It introduces task tokenization to represent different task specifications and employs a shared Transformer backbone to learn transferable representations from heterogeneous time-series datasets spanning different domains, sampling rates, and temporal scales. The pretrained model can be adapted to downstream tasks such as classification, forecasting, imputation, and anomaly detection. In our implementation, we load the released UniTS x128 pretrained backbone and keep it frozen, using ten fixed zero prompt tokens and masked mean pooling over valid chunks, patches, and channels to obtain 128-dimensional representations for downstream evaluation.

    \item \textbf{MOMENT}. MOMENT is a family of general-purpose time-series foundation models pretrained on the Time Series Pile, a large collection of publicly available time-series datasets from diverse domains. Its pretraining framework is designed to learn transferable temporal representations that can be adapted to a range of downstream tasks under limited supervision, including classification, forecasting, imputation, and anomaly detection. We load the pretrained MOMENT-base checkpoint through the official package in embedding mode, freeze the encoder, and apply masked mean pooling over valid chunks, patches, and channels to obtain 768-dimensional representations for each 30-second epoch.

    \item \textbf{Zeus}. Zeus is a general-purpose time-series foundation model developed for tuning-free adaptation across heterogeneous analysis tasks. It combines point-wise tokenization with a multi-scale Transformer organized in a U-shaped hierarchy and introduces multi-objective temporal masking to capture temporal information required by different tasks within a shared pretrained model. This design enables the same model to perform multiple downstream tasks without task-specific fine-tuning. We use the official pretrained checkpoint as a frozen feature extractor, concatenating temporally pooled features from its five hidden-state scales and averaging across signal chunks and channels under our downstream protocol.

    \item \textbf{LaBraM}. LaBraM is an EEG foundation model designed to learn transferable representations across heterogeneous EEG datasets. It partitions multichannel EEG signals into channel-wise patches and uses a vector-quantized neural tokenizer to convert continuous EEG segments into discrete neural codes, after which a Transformer is pretrained to recover the codes of masked EEG patches. Because LaBraM is specifically designed for EEG rather than complete PSG recordings, we include it only for downstream tasks in which the required EEG inputs are available.

    \item \textbf{SleepFM}. SleepFM is a multimodal sleep foundation model pretrained on large-scale PSG recordings spanning EEG, ECG, EMG, and respiratory signals. It adopts a channel-agnostic architecture together with leave-one-out contrastive learning to learn across heterogeneous PSG configurations, encouraging representations from one physiological modality to align with information from the remaining modalities. The pretrained representations are designed to support transfer across sleep-related and clinical prediction tasks. We load the released SleepFM checkpoint with its accompanying model configuration and freeze the encoder, applying its spatial and temporal attention pooling within each modality and concatenating the resulting modality representations into a 512-dimensional vector.

    \item \textbf{SleepMaMi}. SleepMaMi is a sleep foundation model that jointly captures short-term physiological patterns and long-range sleep structure through a hierarchical dual-encoder architecture. Its Micro-Encoder models fine-grained biosignal characteristics using masked autoencoding and multimodal contrastive learning, while its Macro-Encoder captures longer-term sleep dynamics and incorporates demographic-guided contrastive learning. This hierarchical design provides representations at both local physiological and extended temporal scales. We use the official pretrained checkpoint with frozen parameters, following its channel and filtering conventions while adapting inference to complete nights and concatenating macro- and micro-encoder features into 3328-dimensional epoch representations.
    
\end{itemize}    

\subsection{Implementation}

SemPSG is pretrained on 8 $\times$ Ascend NPUs using AdamW with a learning rate of $10^{-4}$ and a global batch size of 64. The pretrained encoder is frozen for downstream evaluation, where task-specific linear heads are trained.

\textbf{Input processing}. PSG recordings are processed as 30-s windows at 128 Hz, yielding 3,840 samples per channel. Continuous signals are normalized per channel at the recording level, without
additional window-level normalization or clipping. Oxygen saturation values are scaled by 0.01. Each waveform is divided into patches of 64 samples, resulting in 60 temporal tokens per channel. Channels are organized into five modality groups: EEG, EOG, EMG, ECG, and RESP. Validity masks identify missing channels and modalities.

\textbf{Model configuration}. The intra-modality and inter-modality encoders each use two Transformer layers with a hidden dimension of 128, four attention heads, a feed-forward dimension of 2,048,
and a dropout rate of 0.1. The separately pretrained channel-name encoder remains frozen throughout PSG pretraining, while the semantic adapter is trainable. The image branch receives three grayscale views of size $160 \times 160$, each arranged into five modality strips of height 32. The CWT view uses real Morlet wavelets with 32 geometrically spaced scales from 1 to 128. The STFT view uses a 256-sample Hann window with a hop size of 64, and the morphology view uses 80 temporal bins. Images are generated offline. The image encoder uses view-specific stems of width 64, $16 \times 16$ patches, and a shared two-layer Transformer with the same hidden dimension, attention head count, feed-forward dimension, and dropout as the waveform encoders. The final window representation consists of six 128-dimensional tokens, one for each physiological modality and one for the image branch. Invalid output tokens are zeroed, and their validity masks are retained.

\textbf{Pretraining}. SemPSG is pretrained on eight Ascend NPUs in FP32 using
AdamW with a learning rate of $10^{-4}$ and weight decay
of 0.01.
The batch size is eight windows per device, giving a
global batch size of 64 without gradient accumulation.
We use 2,000 warmup steps followed by cosine learning-rate
decay and clip the gradient norm at 1.0.
Training runs for 80,245 optimizer steps with random seed 42.
The masking probabilities for channel-level temporal tokens,
modality-level temporal tokens, and image patches are
0.35, 0.25, and 0.40, respectively. Masking is applied only to valid tokens.
The contrastive temperature is 0.07.
Validation is performed every 5,000 steps using at most
20 global batches of 64 windows. For checkpoint selection, the validation weights for
intra-modality reconstruction, inter-modality reconstruction,
image reconstruction, cross-view contrast, and
waveform-image alignment are
$(1,1,\lambda_{\mathrm{mim}},1,1)$, respectively.

\textbf{Downstream adaptation}. The entire pretrained encoder is frozen during downstream
evaluation, and only task-specific prediction heads are
trained on cached representations.
For window-level linear probes, the six output tokens
are concatenated into a 768-dimensional feature vector.

\subsection{Additional Experimental Results}

\begin{figure}[t]
    \centering
    \includegraphics[width=0.7\linewidth]{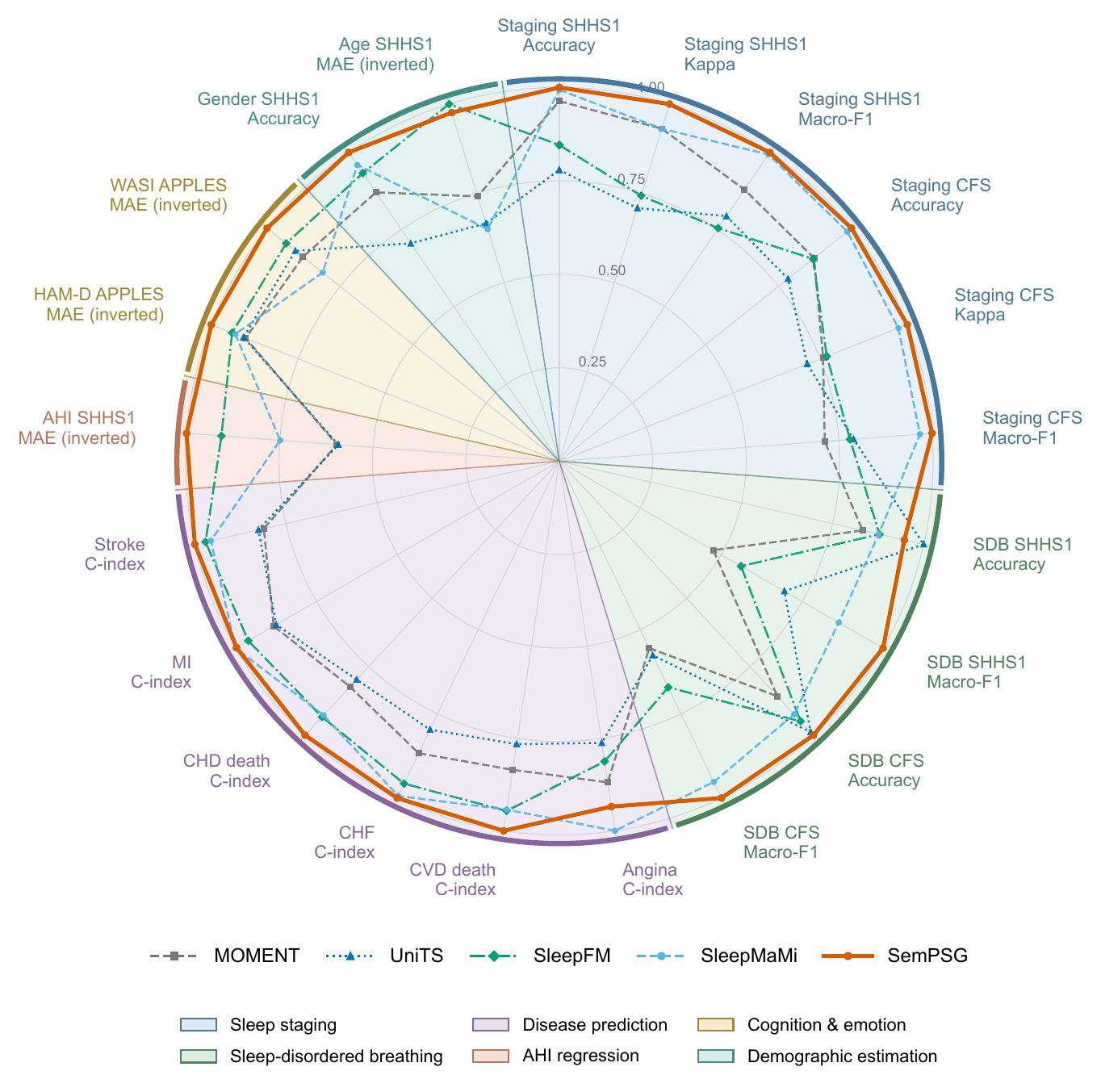}
\caption{\textbf{Overall downstream performance of SemPSG and competing foundation models.}
Each axis represents a normalized evaluation metric from one of six downstream task groups. MAE-based metrics are inverted such that higher values consistently indicate better performance. Colored sectors denote different task categories.}
\label{fig:radar_overview}
\end{figure}

Figure~\ref{fig:radar_overview} provides an overall comparison across six downstream task groups, including sleep staging, sleep-disordered breathing analysis, AHI regression, disease prediction, cognition and emotion assessment, and demographic estimation. For visualization, all metrics are normalized to a common scale, with MAE-based metrics inverted so that larger values consistently indicate better performance. SemPSG exhibits strong and balanced performance across most task categories, with its curve remaining close to the outer boundary of the radar chart. In contrast, competing models tend to perform strongly on particular tasks but show larger variations across different physiological and clinical outcomes. This comparison highlights the broad transferability of SemPSG representations across heterogeneous downstream settings.%

\subsubsection{Sleep-Disordered Breathing Analysis: AHI Regression}
\label{a:ahi}
Another type of sleep-disordered breathing analysis is apnea–hypopnea index (AHI) regression, SemPSG achieves a 9.4\% lower MAE than the second-best model SleepFM, indicating more accurate estimation of recording-level respiratory disturbance severity. These results show that the learned PSG representations capture information relevant to both fine-grained breathing abnormalities and overall respiratory burden. By preserving modality-specific representations while modeling intra-modality channel dependencies and inter-modality interactions, SemPSG captures respiratory patterns alongside physiological responses distributed across other PSG signals, supporting both second-by-second SDB classification and recording-level AHI estimation.

\begin{table}[t]
\centering
\caption{AHI regression performance on SHHS1.}
\label{tab:ahi_shhs1}
\setlength{\tabcolsep}{4pt}
\renewcommand{\arraystretch}{0.95}

\begin{tabular*}{0.55\linewidth}{
@{\extracolsep{\fill}}lcc@{}
}
\toprule
\textbf{Category} & \textbf{Model} & \textbf{MAE $\downarrow$} \\
\midrule
\multirow{2}{*}{\shortstack[l]{General-purpose\\TSFM}}
& MOMENT & 10.54 \\
& UniTS & 10.61 \\
\cmidrule(lr){1-3}
\multirow{3}{*}{\shortstack[l]{Sleep Foundation\\Model}}
& SleepFM & \underline{6.94} \\
& SleepMaMi & 8.40 \\
& SemPSG & \textbf{6.29} \\
\bottomrule
\end{tabular*}

\end{table}

\subsubsection{Demographic Estimation}
\label{a:de}
We evaluate whether the learned PSG representations encode demographic information through gender classification and age regression. As shown in Table 6, SemPSG achieves the highest accuracy for gender classification and the second-lowest MAE for age estimation. The results indicate that the learned representations retain information associated with demographic characteristics such as gender and age.

\begin{table*}[t]
\centering
\caption{Demographic estimation performance on SHHS1.}
\label{tab:demographic}

\begin{tabular*}{0.98\textwidth}{
@{\extracolsep{\fill}}
c c l c c
@{}
}
\toprule
\textbf{Dataset} & \textbf{Category} & \textbf{Model}
& \textbf{Gender Accuracy $\uparrow$}
& \textbf{Age MAE $\downarrow$} \\
\midrule

\multirow{5}{*}{SHHS1}
& \multirow{2}{*}{General-purpose TSFM}
& MOMENT & 72.7 & 8.28 \\
& & UniTS & 58.9 & 9.22 \\
\cmidrule(lr){2-5}

& \multirow{3}{*}{Sleep Foundation Model}
& SleepFM & 77.9 & \textbf{6.14} \\
& & SleepMaMi$^*$
& \underline{80.1} & 9.45 \\
& & SemPSG
& \textbf{83.5} & \underline{6.29} \\

\bottomrule
\end{tabular*}
\vspace{2pt}
\begin{flushleft}
\footnotesize
$^*$ SleepMaMi is evaluated using only its Micro-Encoder for both gender classification and age regression, as its Macro-Encoder is pretrained with the demographic contrastive objective.
\end{flushleft}
\end{table*}

\begingroup
\raggedbottom
\setlength{\intextsep}{8pt plus 2pt minus 2pt}
\setlength{\abovecaptionskip}{5pt}
\subsubsection{Sleep Staging}
\label{a:ss}

Figure~\ref{fig:confusion} provides a stage-wise analysis of SemPSG’s sleep-staging predictions. Both datasets exhibit predominantly diagonal confusion matrices, with recall exceeding 76\% for Wake, N2, N3, and REM. Wake achieves the highest recall on SHHS-1 and CFS (87.7\% and 90.8\%, respectively), while N2 and REM maintain recall above 80\% on both datasets. N3 recall is 76.7\% on SHHS-1 and 81.0\% on CFS, with most N3 errors assigned to N2 (23.1\% and 18.0\%, respectively). The reverse confusion is substantially smaller: only 5.8\% and 6.3\% of true N2 epochs are classified as N3.
\begin{figure}[H]
    \centering
    \includegraphics[width=0.82\linewidth]{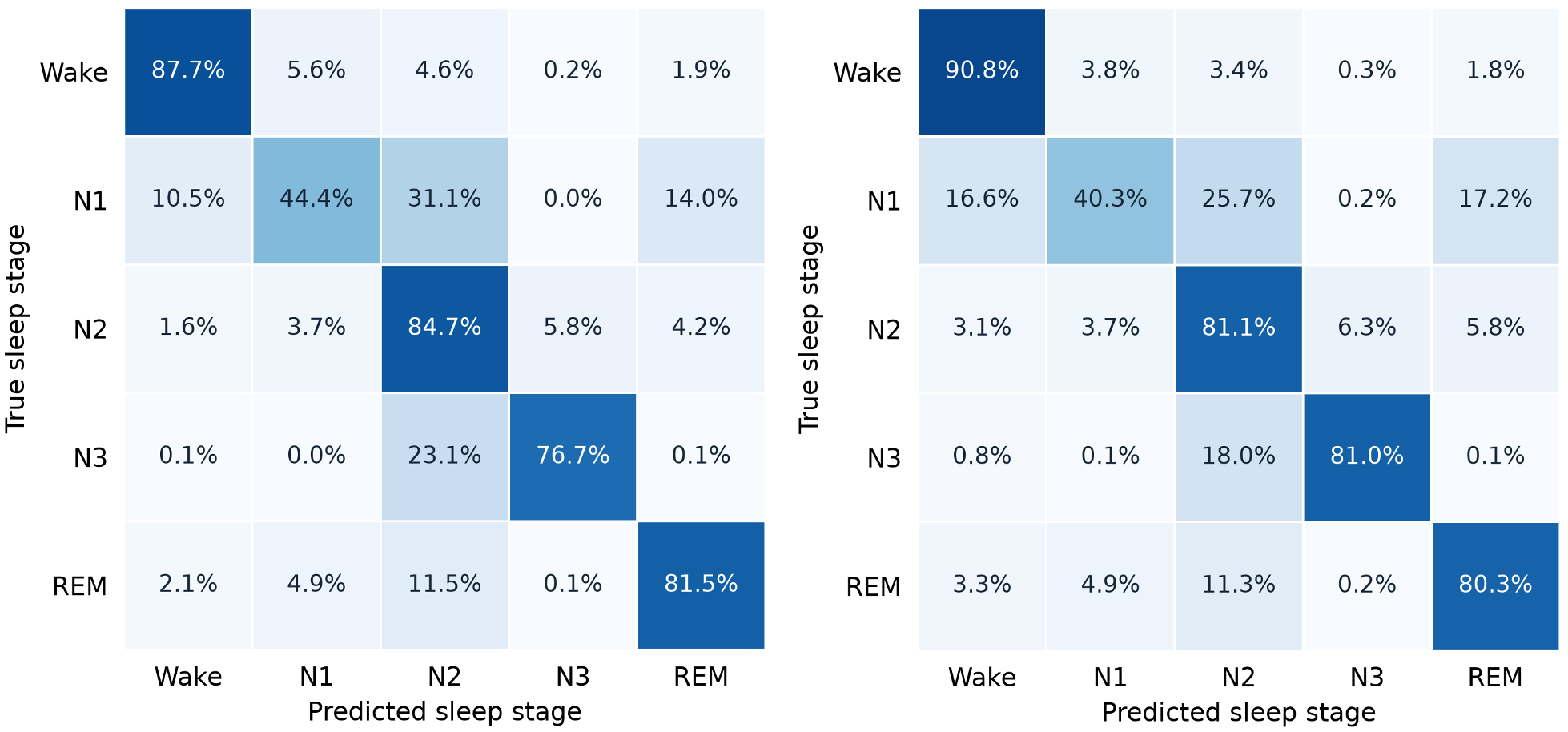}
    \caption{Row-normalized confusion matrices for sleep staging using frozen SemPSG representations with validation-selected linear probes on the SHHS-1 (left) and CFS (right) test sets. Rows indicate ground-truth stages and columns indicate predicted stages. Each entry reports the percentage of epochs within the corresponding ground-truth stage; diagonal entries therefore represent per-stage recall.}
    \label{fig:confusion}
\end{figure}

\clearpage
\subsection{Channel Semantics and Modality Analysis}

Figure~\ref{fig:cne_cosine_sim} and figure~\ref{fig:channel_gate} extend the semantic visualizations in the main text to all five modality groups. Figure~\ref{fig:cne_cosine_sim} presents pairwise cosine similarities between learned channel-name embeddings. The matrix exhibits modality-related structure without forming uniformly similar blocks. ECG channels show relatively high within-group similarity, whereas EEG, EOG, and EMG display more heterogeneous patterns, with stronger similarities among particular channel subsets. Respiratory-related channels also form distinct subgroups rather than a single homogeneous block. These observations indicate that the learned embeddings capture distinctions within the broad modality categories.

Figure~\ref{fig:channel_gate} examines how channel-name assignments affect channel aggregation while keeping the waveform inputs fixed. Shuffling the names changes the relative gating weights of several channels, most visibly within EEG, EMG, and the respiratory-related group. For example, the contrast between the EEG channel rows changes, and the relative weighting of the EMG and leg channels is redistributed. EOG and ECG show comparatively small visual changes in this window, indicating that sensitivity to name reassignment varies across modality groups. The differences primarily concern channel-wise weighting, while the weights remain relatively stable over time within the displayed segment.

Together, these visualizations connect the structure of the channel-name embedding space with the model’s use of channel metadata during aggregation. The shuffling comparison demonstrates that channel weighting depends on the assigned names for the illustrated input. It does not, by itself, establish the contribution of this dependence to downstream performance; the gating weights characterize model behavior rather than physiological channel importance.

\begin{figure}[H]
    \centering
    \includegraphics[width=0.76\linewidth]{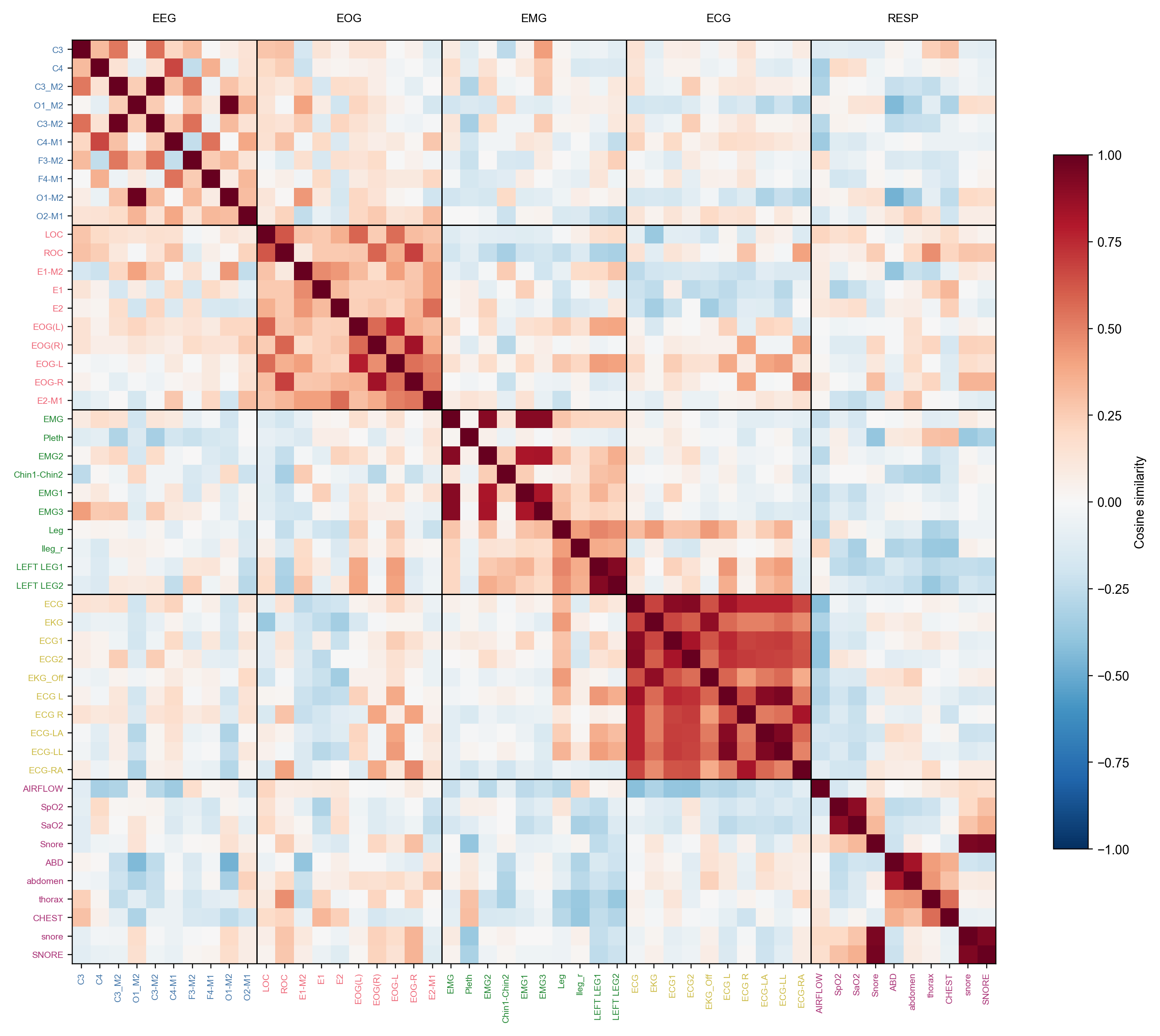}
    \caption{Full pairwise cosine similarity matrix of learned channel-name embeddings across EEG, EOG, EMG, ECG, and respiratory-related channels, extending the subset shown in the main text.}
    \label{fig:cne_cosine_sim}
\end{figure}

\begin{figure}[H]
    \centering
    \includegraphics[width=0.64\linewidth]{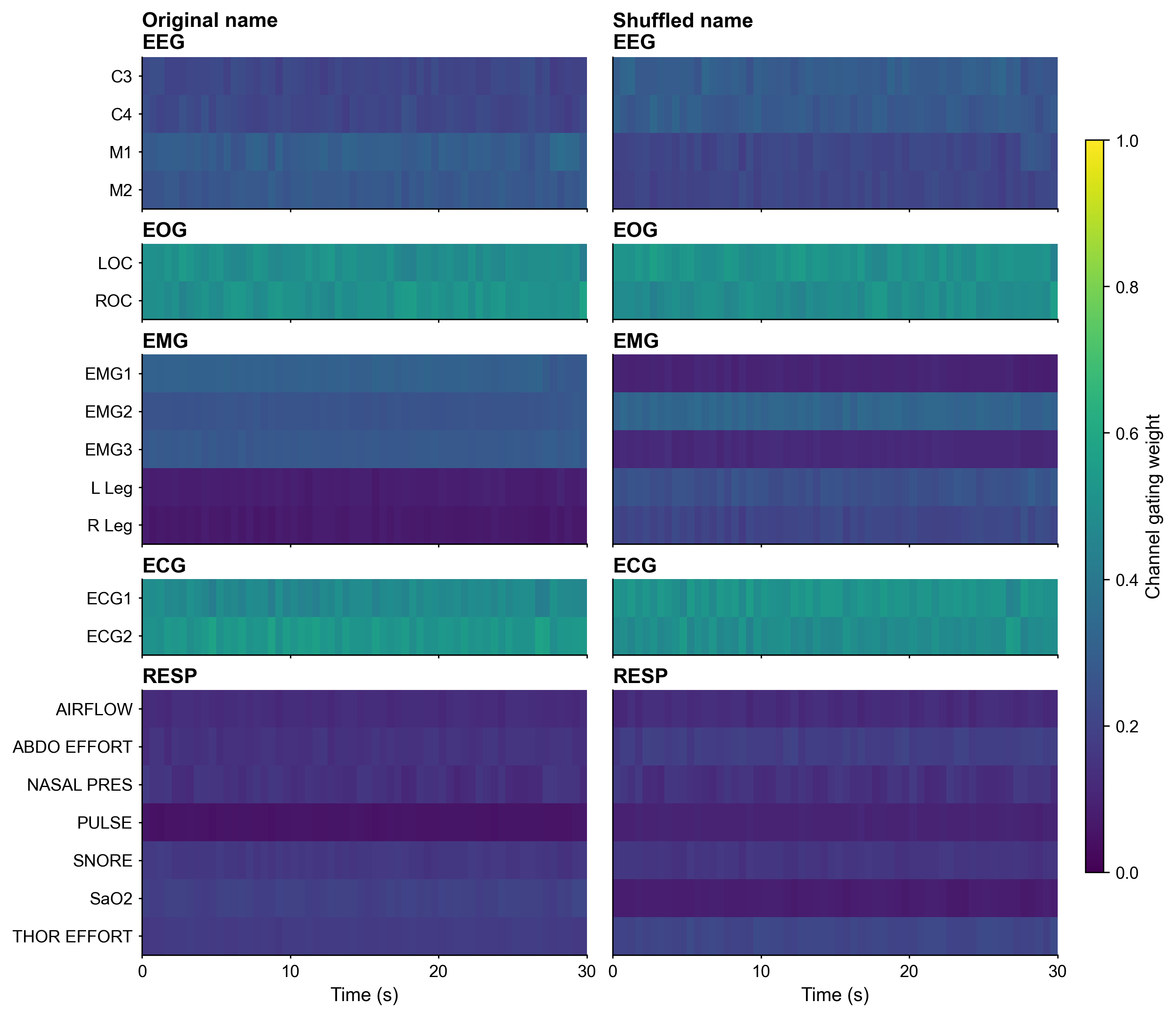}
    \caption{Learned channel-gating weights across all five modality groups over a 30-s window under the original (left) and shuffled (right) channel-name assignments, extending the visualization in the main text.}
    \label{fig:channel_gate}
\end{figure}

\subsection{Analysis of Pretrained Representations}
\label{a:apr}

Figure~\ref{fig:encoder_emb_umap} visualizes the representations extracted from three pretrained models with their parameters frozen. SleepMaMi and SleepFM exhibit substantial overlap among sleep stages, although some regions show stage-specific concentrations. In SleepMaMi, Wake epochs extend toward the right of the main group, while the remaining stages are largely intermingled. SleepFM produces a more elongated distribution, with local concentrations of Wake and N3 but considerable overlap throughout the projection.

SemPSG exhibits a more visible stage-related organization: Wake epochs concentrate toward the upper region, N3 epochs are more prevalent in the lower region, and N2 occupies much of the intervening space. N1 and REM remain interspersed with other stages, indicating that the representation does not yield fully separated stage clusters. This pattern suggests that SemPSG captures sleep-stage structure before downstream task-specific training, while retaining overlap between stages.

\begin{figure}[H]
    \centering
    \includegraphics[width=0.95\linewidth]{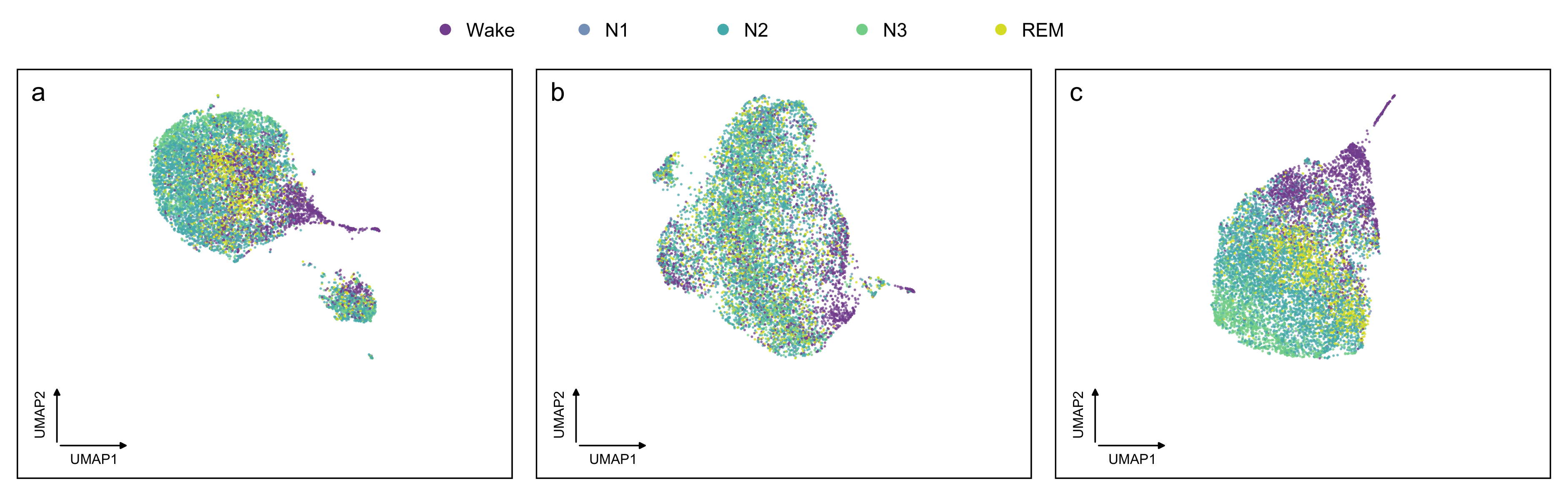}
    \caption{UMAP visualizations of frozen pretrained representations from (a) SleepMaMi, (b) SleepFM, and (c) SemPSG on SHHS-1. Each point represents a sleep epoch and is colored by its ground-truth sleep stage: Wake, N1, N2, N3, or REM. The projections provide a qualitative view of how sleep-stage information is organized in the pretrained representation spaces.}
    \label{fig:encoder_emb_umap}
\end{figure}

\par
\endgroup
\end{document}